\documentclass{article}

 \usepackage[preprint]{neurips_2026}

\usepackage[utf8]{inputenc} 
\usepackage[T1]{fontenc}    
\usepackage{hyperref}       
\usepackage{url}            
\usepackage{booktabs}       
\usepackage{amsfonts}       
\usepackage{nicefrac}       
\usepackage{microtype}      
\usepackage{xcolor}         
\usepackage{multirow}
\usepackage{graphicx}
\usepackage{subcaption}
\usepackage{wrapfig}
\usepackage{longtable}
\usepackage{array}
\usepackage{tabularx}
\usepackage{enumitem}
\usepackage{placeins}
\usepackage[table]{xcolor}
\definecolor{groupgray}{RGB}{235,235,235}
\hypersetup{hidelinks}

\title{Video-IFBench: Evaluating Instruction Following of Multimodal LLMs in Video Understanding Scenarios}

\author{%
  \textbf{Hongbo Liu}\textsuperscript{1,*} \, \textbf{Peixian Chen}\textsuperscript{2,*} \, \textbf{Sihan Liu}\textsuperscript{2} \, \textbf{Peiyuan Zhang}\textsuperscript{3} \, \textbf{Kai Zou}\textsuperscript{4} \\
  \textbf{Dian Zheng}\textsuperscript{5} \, \textbf{Xiaoxing Hu}\textsuperscript{3} \, \textbf{Yuhao Dong}\textsuperscript{6} \, \textbf{Mengdan Zhang}\textsuperscript{2} \, \textbf{Yunhang Shen}\textsuperscript{2} \\
  \textbf{Haoyu Cao}\textsuperscript{2} \, \textbf{Wei Liu}\textsuperscript{2} \, \textbf{Weibo Gu}\textsuperscript{2} \, \textbf{Xing Sun}\textsuperscript{2} \, \textbf{Shengjie Zhao}\textsuperscript{1}\textsuperscript{$\dagger$} \\
  \textsuperscript{1}TJU \quad \textsuperscript{2}Tencent Youtu Lab \quad \textsuperscript{3}SJTU \quad \textsuperscript{4}Tencent Hunyuan \quad \textsuperscript{5}CUHK \quad \textsuperscript{6}NTU \\
  \textsuperscript{*}Equal contribution. \quad \textsuperscript{$\dagger$}Corresponding author.\\
  Project page: \href{https://alexios-hub.github.io/Video-IFBench/}{\texttt{https://alexios-hub.github.io/Video-IFBench/}}
}

\begin{document}

\maketitle

\begin{abstract}
Multimodal Large Language Models (MLLMs) have shown strong performance in video understanding. However, their ability to follow instructions in this domain remains under-explored. Real-world video understanding requires models not only to interpret video content correctly, but also to satisfy diverse user-specified constraints. Existing benchmarks focus primarily on task accuracy rather than instruction adherence, leaving this capability insufficiently evaluated. To address this gap, we introduce Video-IFBench, a comprehensive benchmark for evaluating instruction following in video understanding, where models must satisfy diverse user-specified constraints, including those grounded in visual and audio content. We develop an instruction taxonomy with four templates, including single-task, multi-task, selection, and nested instructions, covering 32 task types and 39 manually designed constraint categories spanning both semantic and format requirements. To reduce annotation cost, we build a semi-automatic data construction pipeline that combines MLLMs, programmatic processing, and human verification, resulting in 1.5K samples. We conduct a large-scale evaluation of more than 20 recent MLLMs and show that video instruction following remains challenging for current models, especially for instructions with many constraints, semantic constraints, or complex conditional structures that require selecting the correct branch or path based on video content. We hope our work will facilitate future research on instruction following in video understanding scenarios.\end{abstract}

\section{Introduction}

Recent Multimodal Large Language Models (MLLMs)~\cite{team2023gemini,zhu2025internvl3,liu2025ola,bai2025qwen3vl,team2026qwen3_5omni,xu2025qwen3omni} have advanced rapidly and achieved strong performance on various video understanding tasks. At the same time, the evaluation of video understanding is rapidly expanding. Existing benchmarks examine diverse capabilities, including temporal understanding~\cite{liu2024tempcompass}, spatial intelligence~\cite{yang2025thinkinginspace}, general video analysis~\cite{fu2025videomme,fu2026videommev2,li2024mvbench,zhao2025mmvu}, long-video comprehension~\cite{zhou2025mlvu,wu2024longvideobench}, and knowledge-intensive reasoning~\cite{hu2025videommmu}. 
However, evaluating video understanding solely through response correctness is increasingly insufficient for real-world applications.

An important yet underexplored dimension is instruction following. In practical video understanding scenarios, user requests are often accompanied by explicit requirements related to semantic~\cite{li2026perceptioncomp,wang2026hopchain,shen2026mmcondchain} and format constraints~\cite{zhou2023IFEval,ding2025mmif,pyatkin2025IFBench} on model responses. They may also vary in structural complexity, from requests that seek a single piece of information from the video to multi-aspect requests and conditional instructions: models must identify the correct condition branch or path based on the actual video content, execute the corresponding instructions, and satisfy the associated constraints. 
For example, when analyzing a cooking tutorial, a user may ask an MLLM to describe only the steps after the ingredients are mixed, and list them in chronological order with timestamps. Therefore, the ability to follow instructions is a critical dimension for assessing whether such models can be reliably deployed in real-world scenarios.

Despite its practical importance, this capability is not yet adequately evaluated. Existing video benchmarks focus primarily on response correctness~\cite{li2024mvbench,fu2025videomme,liu2024tempcompass,zhou2025mlvu,wu2024longvideobench,fu2026videommev2}. 
They are not designed to test whether MLLMs faithfully execute complex user instructions. Meanwhile, prior work on instruction-following evaluation focuses mainly on text-only settings~\cite{jing2023followeval,jiang2024followbench,zhou2023IFEval,qin2024infobench,wen2024complexbench,pyatkin2025IFBench}, or on relatively narrow multimodal settings such as image-based instruction following~\cite{ding2025mmif,qian2024mia,he2026vcifeval}, and controllable video captioning~\cite{li2025ifvidcap}. As a result, instruction following in more diverse forms, especially under richer visual/audio-grounded constraints, remains insufficiently explored.


To address this gap, we introduce \textbf{Video-IFBench}, a comprehensive benchmark for evaluating instruction-following capabilities in video understanding. Video-IFBench contains over $700$ videos from public data sources, covering diverse domains with individual durations ranging from $10$ seconds to $10$ minutes and a total duration of approximately $49$ hours. We design an instruction taxonomy with four templates, including single-task, multi-task, selection, and nested instructions, covering $32$ manually curated video understanding task types and $39$ semantic or format response constraints. To reduce annotation cost, we develop a semi-automatic data construction pipeline that combines MLLM-powered extraction of global and fine-grained local video information, multi-stage complex instruction and checklist generation driven by both MLLMs and programmatic rules, and rigorous human verification, resulting in $1.5$K high-quality samples.

Beyond data construction, we design an evaluation protocol that combines LLM-as-Judge with programmatic verification, and conduct a large-scale empirical study on more than $20$ recent MLLMs. Our results show that even the strongest video MLLMs in our evaluation still leave substantial room for improvement in instruction following, with the best model achieving only $54.5$\% overall score. We further find that models struggle particularly with instructions that involve many constraints, semantic constraints, and complex conditional structures that require matching the correct branch or path based on video content. These findings suggest that strong video understanding accuracy does not automatically translate into faithful adherence to user intent. We therefore argue that instruction following should be treated as a first-class evaluation dimension for video MLLMs.

In summary, this work makes the following contributions:  
\begin{itemize}[left=0pt]
    \item We introduce \textbf{Video-IFBench}, a new benchmark for video instruction following, containing 1.5K high-quality samples over more than 700 videos with diverse instruction structures, tasks, and constraints.

    \item We design a hybrid evaluation protocol that combines LLM-as-Judge with programmatic verification to assess models performance on video instruction following.

    \item We evaluate more than 20 recent MLLMs and show that current models still struggle with strict video instruction following, especially under many constraints, semantic constraints, and conditional instructions.
\end{itemize}

\section{Related Work}

\subsection{Instruction-Following Evaluation on LLMs.}
Instruction-following evaluation has been widely studied in text-only settings. Benchmarks such as FollowEval~\cite{jing2023followeval}, FollowBench~\cite{jiang2024followbench}, IFEval~\cite{zhou2023IFEval}, and InFoBench~\cite{qin2024infobench} evaluate whether models satisfy explicit and structured instruction requirements. Later work further studies more complex and realistic settings, including compositional constraints~\cite{wen2024complexbench}, real-user multi-constraint instructions~\cite{lior2025wildifeval}, unseen constraints~\cite{pyatkin2025IFBench}, sequential or multi-turn instructions~\cite{chen2024sifo,he2024multi-if,li2025structflowbench}, and agentic tasks~\cite{qi2025agentif,chen2026agentifoneday}.

\subsection{Multimodal Instruction Following.}
Recent benchmarks extend instruction-following evaluation to multimodal inputs. MIA-Bench~\cite{qian2024mia}, MM-IFEngine~\cite{ding2025mmif}, VC-IFEval~\cite{he2026vcifeval}, MMMT-IF~\cite{epstein2024mmmt}, and MCIF~\cite{papi2025mcif} evaluate instruction following across image, speech, dialogue, and crosslingual settings. In the video domain, IF-VidCap~\cite{li2025ifvidcap} studies constrained video captioning. In contrast, our work evaluates instruction following across broader video understanding tasks and more complex instruction structures.

\subsection{Video Understanding Benchmarks.}
Existing video understanding benchmarks evaluate temporal understanding, long-video comprehension, knowledge-intensive reasoning, robustness, and streaming interaction, including MVBench~\cite{li2024mvbench}, Video-MME~\cite{fu2025videomme}, TempCompass~\cite{liu2024tempcompass}, Video-MMMU~\cite{hu2025videommmu}, MLVU~\cite{zhou2025mlvu}, MMVU~\cite{zhao2025mmvu}, LongVideoBench~\cite{wu2024longvideobench}, PerceptionComp~\cite{li2026perceptioncomp}, Video-MME-v2~\cite{fu2026videommev2}, PhoStream~\cite{lu2026phostream}, and OVO-Bench~\cite{niu2025ovo}. These benchmarks primarily measure response correctness, but provide limited evaluation of whether models can faithfully follow complex user instructions with multiple constraints and conditional structures.

\section{Video-IFBench}
\begin{figure}
    \centering
    \makebox[\linewidth][c]{%
        \includegraphics[width=1.0\linewidth]{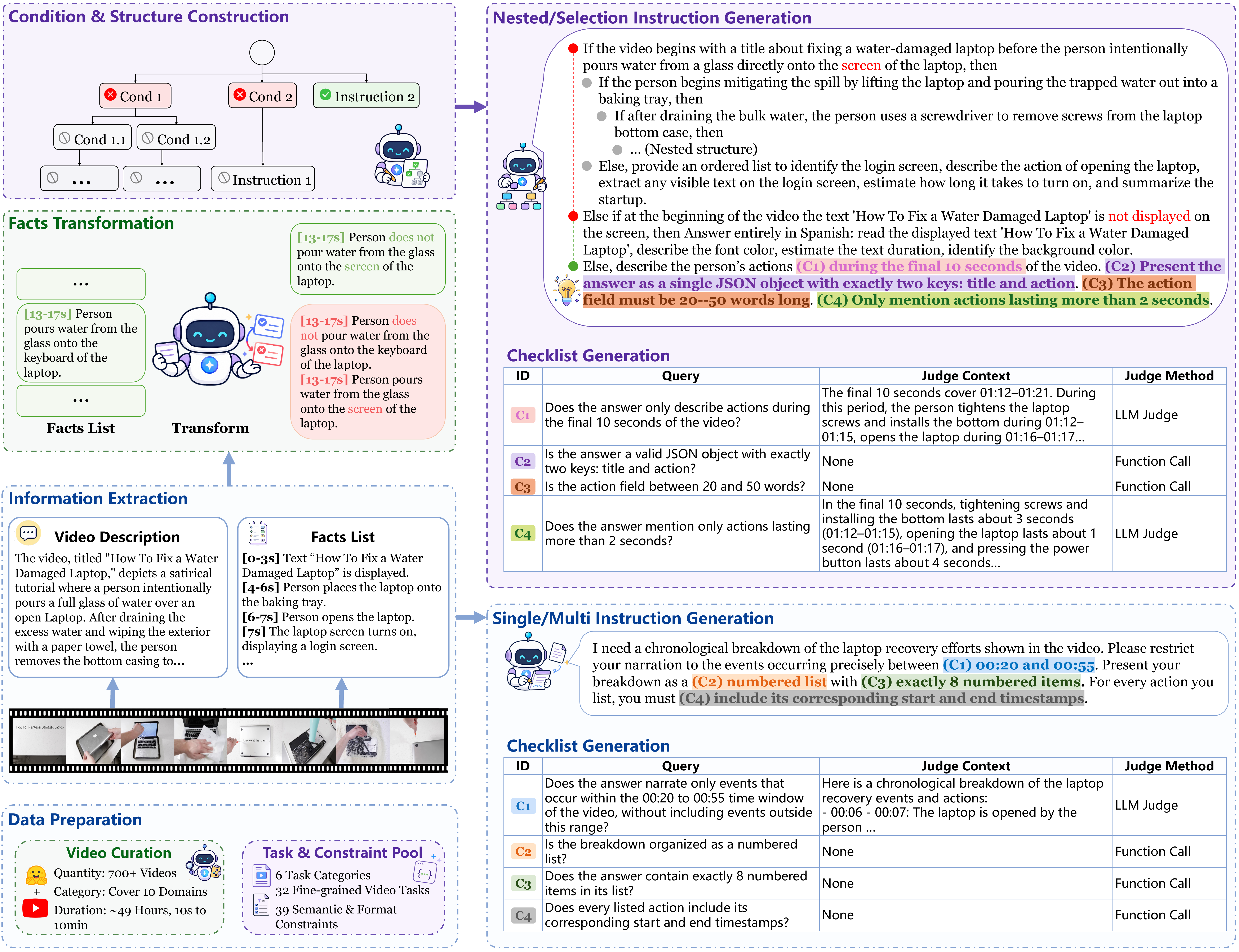}
    }
    \vspace{-1em}
\caption{The data construction pipeline of Video-IFBench. Each curated video is annotated with a global description and fine-grained timestamped facts, which are used to generate Single/Multi instructions and condition-based Selection/Nested instructions. Each instruction is paired with a checklist for LLM-based or programmatic verification. More examples are provided in Appendix~\ref{appendix:example_vis}.}
\label{fig:pipeline}
\end{figure}

\subsection{Benchmark Design}
To comprehensively evaluate the video instruction-following ability of MLLMs, we design an instruction taxonomy with four instruction templates: \textbf{Single}, \textbf{Multi}, \textbf{Selection}, and \textbf{Nested}. These target two key dimensions of instruction complexity: task composition and conditional structure.

\paragraph{Single and Multi.}
Single and Multi instructions vary task-level composition. A Single instruction contains one task that asks for a specific piece of information from the video, together with a set of response constraints. We use this template to study how performance changes as the number of constraints increases, with constraint counts ranging from $1$ to $15$. In contrast, a Multi instruction contains two or more tasks and requires the model to retrieve or reason over multiple pieces of video information in a single request. Its constraints may apply either globally to the entire response or locally to specific tasks. To focus on the effect of task count, we keep the number of global and task-specific constraints relatively small, mostly no more than three.

\paragraph{Selection and Nested.}
Selection and Nested instructions evaluate whether models can follow video-grounded conditions. In Selection, the model chooses the correct conditional branch from several candidate branches and executes the corresponding instruction. Nested further organizes such conditions into a tree structure, requiring the model to follow the correct path before answering.
\begin{figure}[t]
    \centering

    \begin{subfigure}[t]{0.25\linewidth}
        \centering
        \includegraphics[height=0.15\textheight,keepaspectratio]{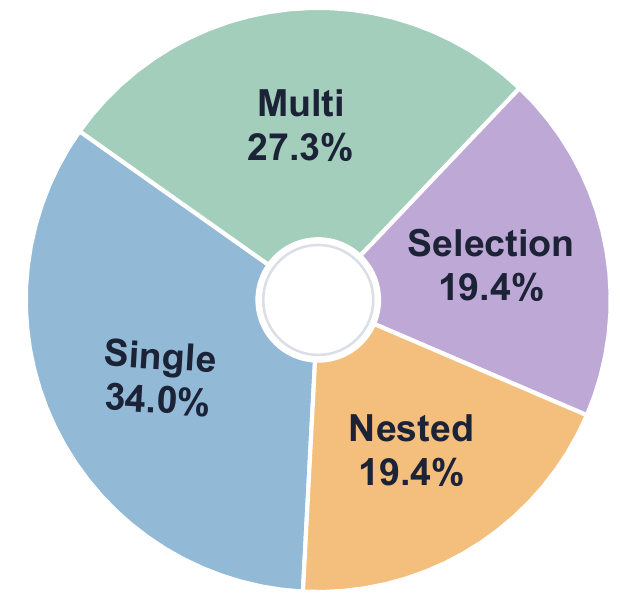}
        \caption{Instruction types.}
        \label{fig:instruction_type_distribution}
    \end{subfigure}
    \hfill
    \begin{subfigure}[t]{0.70\linewidth}
        \centering
        \includegraphics[height=0.15\textheight,keepaspectratio]{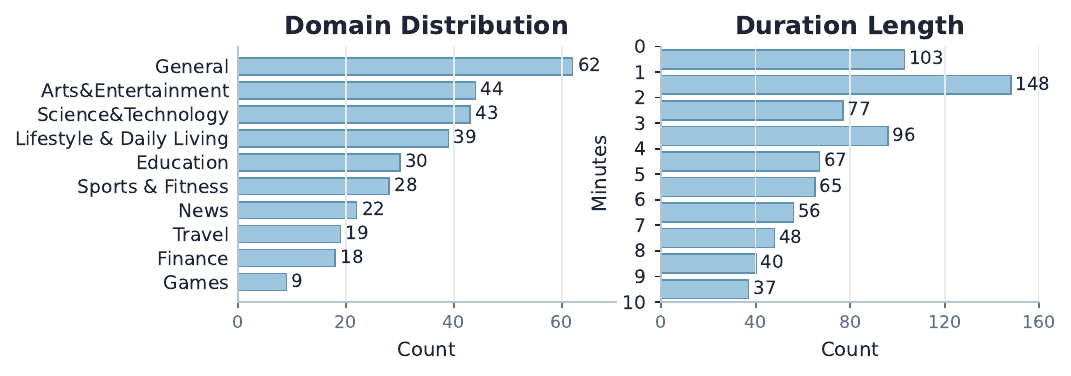}
        \caption{Video domains and durations.}
        \label{fig:domain_duration_distribution}
    \end{subfigure}

    \vspace{-0.4em}
    \caption{Dataset-level statistics of Video-IFBench.}
    \label{fig:dataset_level_statistics}
    \vspace{-1em}
\end{figure}

\subsection{Dataset Construction}

As shown in Fig.~\ref{fig:pipeline}, our pipeline includes video curation, task and constraint pool construction, information extraction, instruction generation, checklist generation, and human verification. We collect and filter public videos, build task and constraint pools, and extract a global description and fine-grained timestamped facts for each video. These annotations are then used to generate Single/Multi instructions and condition-based Selection/Nested instructions. We further generate checklists for automatic evaluation and manually verify the constructed samples.
\paragraph{Video Curation.}
We collect videos from diverse public sources, including existing video understanding benchmarks (WorldSense~\cite{hong2025worldsense}, Video-MME~\cite{fu2025videomme}, MMBench-Video~\cite{fang2024mmbenchvideo}, FineVideo~\cite{Farr2024FineVideo}, and LongVALE~\cite{geng2025longvale}). We also use a set of manually designed keywords to retrieve additional videos from public platforms, covering ten domains. To ensure visual quality, we use MUSIQ~\cite{ke2021musiq} to filter out blurred videos. The domain and duration distributions are shown in Fig.~\ref{fig:domain_duration_distribution}.

\paragraph{Task and Constraint Pool Construction.}
We analyze samples from existing video understanding benchmarks~\cite{fu2025videomme,hong2025worldsense,geng2025longvale,fang2024mmbenchvideo,hu2025videommmu} and group representative tasks into six high-level categories (Perception \& Recognition, Temporal Understanding, Spatial Understanding, Relationship \& Interaction, Logical Reasoning, Captioning). For each category, we manually define multiple concrete task types, each with a task name, a brief description, and example questions. This design improves task diversity over directly prompting MLLMs to generate questions. The full task taxonomy is reported in Appendix~\ref{appendix:task_taxonomy}.

We construct the constraint pool by reusing and extending format constraints from prior instruction-following benchmarks~\cite{zhou2023IFEval,ding2025mmif,pyatkin2025IFBench}, and designing semantic constraints that capture video-specific requirements (\textit{e.g.}, temporal scope, relations). After multiple rounds of manual cleaning to remove ambiguous, redundant, or hard-to-verify constraints, we obtain $39$ constraint types, including $22$ semantic and $17$ format constraints. The full constraint taxonomy is provided in Appendix~\ref{appendix:constraint_taxonomy}.

\paragraph{Information Extraction.}
For each video, we use an MLLM to extract information at two granularities: a global description of the overall content and a fine-grained facts list. Each fact is defined as an atomic, timestamped visual or audio statement grounded in video content.

For longer videos, extracting the full facts list at once often produces coarse and incomplete annotations. We therefore split each long video into one-minute segments and process them sequentially. At each step, we prompt the MLLM to extract local facts and update a shared entity memory. This helps keep references to people and objects consistent across segments.

\paragraph{Single and Multi Instruction Generation.}
We sample one or more tasks according to the target instruction type and prompt the MLLM to generate instructions from the video description, facts list, and sampled constraints. 

To avoid repetitive constraint patterns, we sample constraints from predefined count ranges and limit how often each constraint type can appear, which helps balance both constraint types and counts in the generated data. Examples of Single and Multi instructions are shown in Figs.~\ref{fig:example_1}--\ref{fig:example_4} in Appendix~\ref{appendix:example_vis}.

\paragraph{Selection and Nested Instruction Generation.}
Directly prompting an MLLM to generate Selection and Nested instructions often fails to produce sufficiently complex branching or nested structures. We therefore use a multi-stage pipeline that constructs the conditional structure before generating branch or leaf instructions.

As shown in the top-left of Fig.~\ref{fig:pipeline}, we first transform facts into true and false variants to construct conditions. Then, the MLLM generates three transformations: an \textit{edit} transformation that minimally changes the fact into a false variant (\textit{e.g.}, ``keyboard'' $\rightarrow$ ``screen''), a \textit{negation} transformation that negates the original fact (\textit{e.g.}, ``pours water'' $\rightarrow$ ``does not pour water''), and a combined transformation that edits and negates the fact simultaneously to produce another true variant.

We then generate the instruction structure and sample one leaf as the unique true target. For each non-target leaf path, we randomly mark one intermediate node as false if the path does not already contain a false node. This ensures that each Selection or Nested instruction has exactly one valid branch or root-to-leaf path. The MLLM then writes the condition text for each intermediate node and generates a constrained instruction for each leaf node. Examples of Selection and Nested instructions are shown in Figs.~\ref{fig:example_5} and~\ref{fig:example_6}, respectively.

\begin{figure}[t]
    \centering

    \begin{subfigure}[t]{0.4\linewidth}
        \centering
        \includegraphics[height=0.25\textheight,keepaspectratio]{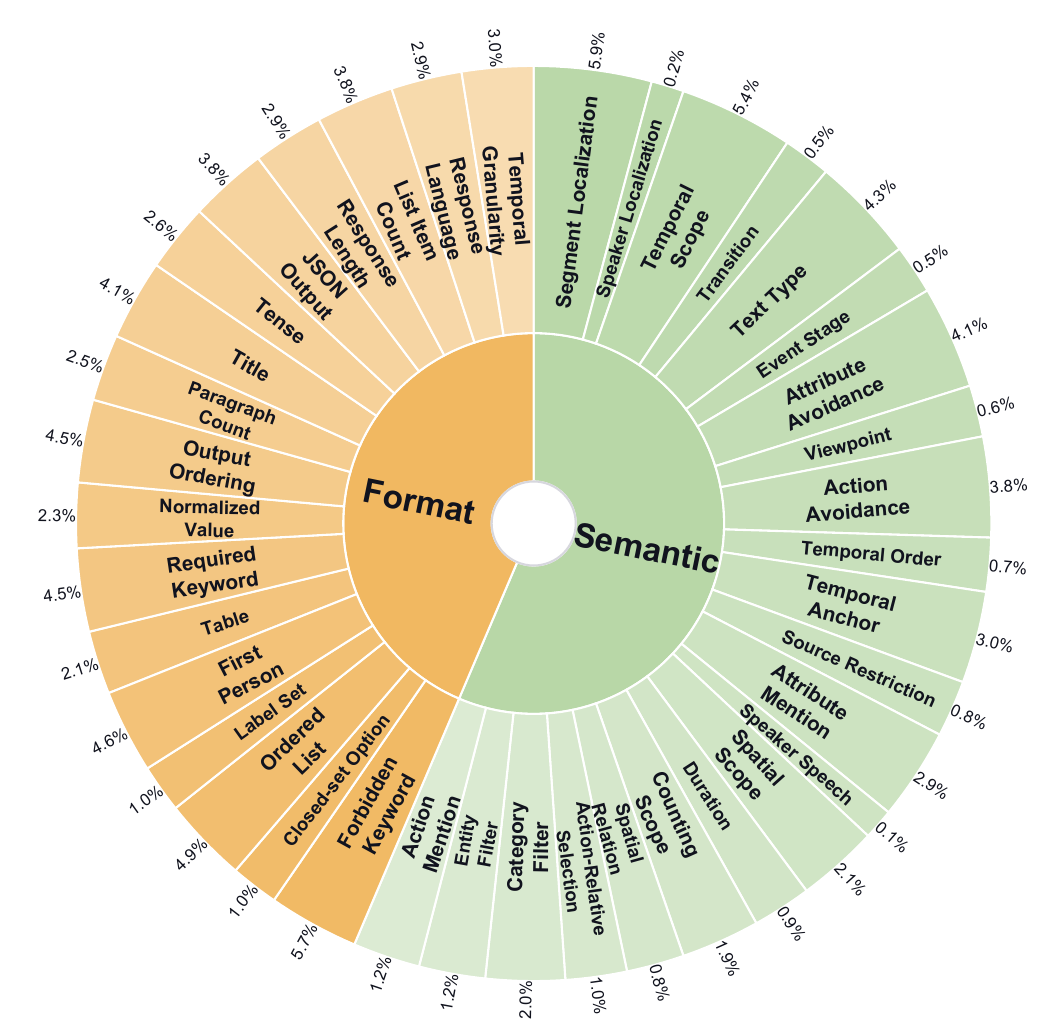}
        \caption{Constraint distribution.}
        \label{fig:constraints_distribution}
    \end{subfigure}
    \hfill
    \begin{subfigure}[t]{0.57\linewidth}
        \centering
        \includegraphics[height=0.2\textheight,keepaspectratio]{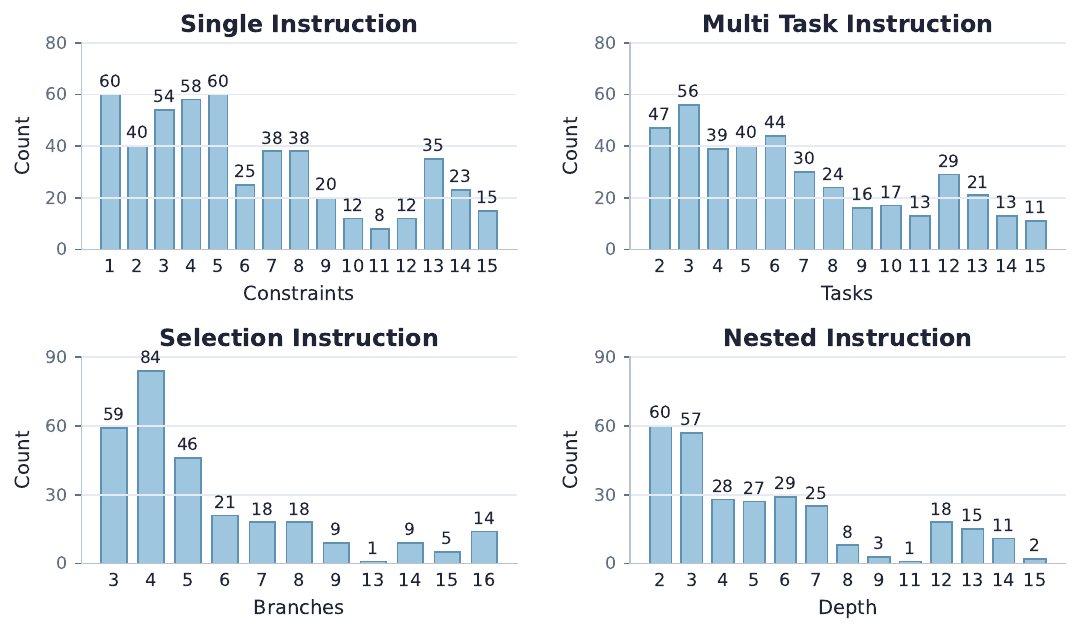}
        \caption{Instruction structure statistics.}
        \label{fig:instruction_structure_distributions}
    \end{subfigure}

    \vspace{-0.4em}
    \caption{Instruction and constraint statistics of Video-IFBench.}
    \label{fig:instruction_constraint_statistics}
    \vspace{-1em}
\end{figure}

\paragraph{Checklist Generation.}
During instruction generation, we record all constraints sampled for each instruction and pair each constraint with a checklist item. Given the instruction and the constraint definition, we prompt the MLLM to generate a binary checklist query, where ``yes'' indicates that the constraint is satisfied. For semantic constraints, we also generate a video-grounded judge context as evidence for evaluation.
\paragraph{Human Verification.}
We manually verify and clean all generated samples. Annotators check video descriptions and facts, instruction constraints, conditional branches or paths, and checklist queries and judge contexts. Problematic samples are corrected or discarded. The final distributions of instruction types, constraints, and instruction structures are shown in Fig.~\ref{fig:instruction_type_distribution}, Fig.~\ref{fig:constraints_distribution}, and Fig.~\ref{fig:instruction_structure_distributions}.

\subsection{Evaluation Paradigm}


We evaluate model responses with a hybrid checklist-based protocol that combines LLM-as-Judge and programmatic verification. Since a sample may contain multiple candidate instructions under conditional structures, we define the \textit{active instruction} as the instruction that should be executed. For Single and Multi samples, the given instruction is directly active. For Selection and Nested samples, only the instruction under the correct branch or path is active.

We then evaluate the responses at two levels. Task-level coverage checks whether the response addresses all required tasks in the active instruction; for conditional instructions, this stage is especially useful because it measures whether the model follows the correct branch or path. Constraint-level adherence is evaluated using the checklist items associated with the active instruction. Each checklist item is judged either by an LLM, using the checklist query and judge context, or by a deterministic verifier for constraints that can be verified programmatically, \textit{e.g.}, length, format, and keywords.

For each sample $i$, let $t_i \in \{0,1\}$ denote task-level coverage, $n_i$ denote the number of checklist items, and $c_{ij} \in \{0,1\}$ denote whether the $j$-th item is satisfied. We report two task-gated metrics:
\begin{equation}
\mathrm{TCSR}=\frac{1}{N}\sum_{i=1}^{N} t_i \cdot \frac{1}{n_i}\sum_{j=1}^{n_i} c_{ij},
\qquad
\mathrm{TISR}=\frac{1}{N}\sum_{i=1}^{N} t_i \cdot \prod_{j=1}^{n_i} c_{ij}.
\end{equation}
TCSR measures average constraint satisfaction after task-level gating, while TISR requires the response to address the active tasks and satisfy all checklist items simultaneously.

\section{Experiments}
\subsection{Settings}
\label{sec:settings}
We conduct an evaluation on frontier proprietary and open-source models, including Gemini-3-Pro\&Flash~\cite{google2025gemini3}, GPT-5.4~\cite{singh2025gpt5}, Doubao~\cite{bytedanceseed2026seed20}, the Gemma 4 series~\cite{google2026gemma4}, the InternVL 3.5 series~\cite{wang2025internvl35}, the Qwen-Omni series~\cite{xu2025qwen25omni,xu2025qwen3omni}, the Qwen3-VL series~\cite{bai2025qwen3vl}, and the Qwen3.5 series~\cite{qwen2026qwen35}. For Gemini and Doubao, we provide videos at 1 FPS; for GPT-5.4, we sample 50 frames due to API constraints. For each open-source model, we follow the recommended inference setup to ensure a representative evaluation. We feed the original video with audio to omni-modal models, and provide vision-only models with video frames and timestamped subtitles transcribed by Gemini-3-Pro. In the main experiments, we use Qwen3.5-397B-A17B-Instruct~\cite{qwen2026qwen35} as the judge model. To examine the effect of judge model choice, we also report results with a smaller Qwen3.5-35B-A3B judge in the Appendix~\ref{appendix:qwen3_5_35b_judge}. The results show that both settings lead to consistent trends. Subsequent analyses are based on the Qwen3.5-397B-A17B-Instruct judge.

\subsection{Main Results}
\label{sec:main_results}
\begin{table*}[t]
\centering
\scriptsize
\caption{Main results on Video-IFBench. We report TCSR and TISR for each instruction type and the overall benchmark. Best results within each model group are in bold.}
\label{tab:main_results}
\resizebox{\textwidth}{!}{
\begin{tabular}{lcccccccccc}
\toprule
\multirow{2}{*}{Model}
& \multicolumn{2}{c}{Single}
& \multicolumn{2}{c}{Multi}
& \multicolumn{2}{c}{Selection}
& \multicolumn{2}{c}{Nested}
& \multicolumn{2}{c}{Overall} \\
\cmidrule(lr){2-3} \cmidrule(lr){4-5} \cmidrule(lr){6-7} \cmidrule(lr){8-9} \cmidrule(lr){10-11}
& TCSR & TISR & TCSR & TISR & TCSR & TISR & TCSR & TISR & TCSR & TISR \\
\midrule

\rowcolor{groupgray}
\multicolumn{11}{c}{\textbf{Proprietary Models}} \\
\midrule
Gemini-3-Pro~\cite{google2025gemini3} & \textbf{79.6}&	\textbf{52.3}&	\textbf{88.6}&	\textbf{58.8}&	\textbf{68.5}&	\textbf{59.2}&	\textbf{53.7}&	\textbf{46}&	\textbf{76.5}&	\textbf{54.5} \\
Gemini-3-Flash~\cite{google2025gemini3} & 76.7 & 49.2 & 87.6 & 56.4 & 63.9 & 54.8 & 39.9 & 30.7 & 72.2 & 49.5 \\
Doubao-Seed-2.0-Pro-260215~\cite{bytedanceseed2026seed20} & 76.7 & 44.6 & 87.1 & 51.1 & 46.5 & 38.2 & 17.8 & 14.4 & 65.7 & 40.8 \\
GPT-5.4~\cite{singh2025gpt5} & 72.8 & 34.7 & 82.4 & 43.2 & 21.1 & 16.9 & 12.1 & 7.6 & 57.4 & 30.0 \\

\midrule
\rowcolor{groupgray}
\multicolumn{11}{c}{\textbf{Open-source Models (Instruct)}} \\
\midrule
Qwen2.5-Omni-3B~\cite{xu2025qwen25omni}      & 35.2&	11.6&	34.0&	8.6&	7.7&	5.4&	6.5&	4.5&	25.8&	8.6 \\
Qwen2.5-Omni-7B~\cite{xu2025qwen25omni}      &46.3&	16.6&	48.6&	14.9&	15.6&	11.5&	8.9&	5.5&	36.0&	13.5 \\
Qwen3-Omni-30B-A3B-Instruct~\cite{xu2025qwen3omni}      & 57.9&	21.4&	62.9&	22.9&	17.2&	14.3&	6.1&	4.1&	44.3&	18.0 \\
Gemma-4-E4B-it~\cite{google2026gemma4}      &56.8&	22.1&	67.5&	29.9&	11.9&	8.2&	6.0&	3.8&	45.1&	19.5 \\
Gemma-4-26B-A4B-it~\cite{google2026gemma4}     & 71.6&	35.7&	79.2&	41.1&	22.7&	19.4&	7.1&	4.5&	55.6&	29.7 \\
Gemma-4-31B-it~\cite{google2026gemma4}     & \textbf{75.8} &	\textbf{43.0}&	82.1&	\textbf{44.4}&	\textbf{29.9}&	\textbf{25.4}&	13.8&	\textbf{10.0}&	\textbf{60.4}&	\textbf{35.4} \\
InternVL3.5-8B-Instruct~\cite{wang2025internvl35}     & 53.4&	19.1&	63.6&	21.8&	16.3&	10.1&	7.6&	3.5&	43.1&	16.0 \\
InternVL3.5-14B-Instruct~\cite{wang2025internvl35}     & 56.2&	19.1&	66.3&	22.6&	18.6&	12.5&	6.2&	3.5&	45.1&	16.6 \\
InternVL3.5-30B-A3B-Instruct~\cite{wang2025internvl35}     & 57.0	& 19.2&	65.9&	24.8&	21.2&	13.8&	\textbf{16.2}&	9.6&	47.4&	18.5 \\
InternVL3.5-38B-Instruct~\cite{wang2025internvl35}     & 61.0&	24.5&	71.8&	27.1&	18.5&	12.7&	14.6&	8.4&	49.8&	20.8 \\
InternVL3.5-241B-A28B-Instruct~\cite{wang2025internvl35}    & 64.5&	24.6&	75.3&	32.2&	16.0&	12.3&	13.6&	10.4&	51.7&	22.7 \\
Qwen3-VL-4B-Instruct~\cite{bai2025qwen3vl}      & 55.6&	19.9&	65.6&	20.9&	15.1&	11.4&	8.7&	5.1&	44.3&	16.4 \\
Qwen3-VL-8B-Instruct~\cite{bai2025qwen3vl}      & 60.1&	23.5&	68.9&	24.7&	17.8&	13.0&	13.5&	8.5&	48.0&	19.6 \\
Qwen3-VL-30B-A3B-Instruct~\cite{bai2025qwen3vl}      & 56.8&	19.9&	70.4&	25.4&	17.0&	12.3&	13.6&	9.5&	47.2&	18.6 \\
Qwen3-VL-235B-A22B-Instruct~\cite{bai2025qwen3vl}      &67.5&	31.9&	78.0&	35.1&	19.5&	14.5&	8.6&	6.0&	53.1&	25.8 \\
Qwen3.5-4B-Instruct~\cite{qwen2026qwen35}      & 54.6&	22.0&	65.5&	23.3&	15.2&	8.8&	7.9&	4.0&	43.9&	17.3 \\
Qwen3.5-9B-Instruct~\cite{qwen2026qwen35}      & 57.8 &	22.6&	70.0&	27.3&	16.2&	12.4&	8.1&	5.0&	46.6&	19.6 \\
Qwen3.5-27B-Instruct~\cite{qwen2026qwen35}      & 66.7 &	29.1&	77.8 &	33.8 &	23.0&	18.0&	11.7&	6.9&	54.0&	25.2 \\
Qwen3.5-35B-A3B-Instruct~\cite{qwen2026qwen35}      & 56.7&	25.7&	75.2&	33.4&	17.5&	12.4&	7.5&	5.0&	48.0&	22.6 \\
Qwen3.5-122B-A10B-Instruct~\cite{qwen2026qwen35}      & 62.7&	29.1&	79.8&	36.8&	20.3&	14.5&	9.5&	5.5&	52.5&	25.3 \\
Qwen3.5-397B-A17B-Instruct~\cite{qwen2026qwen35}      & 70.1&	36.5&	\textbf{83.2}&	40.8&	24.3&	18.9&	12.2&	7.9&	57.3&	30.4 \\

\midrule
\rowcolor{groupgray}
\multicolumn{11}{c}{\textbf{Open-source Models (Thinking)}} \\
\midrule
Qwen3-Omni-30B-A3B-Think~\cite{xu2025qwen3omni}      & 66.6 &	30.0&	75.1&	36.8&	24.8&	18.5&	10.1&	5.5&	53.1&	26.2 \\
Qwen3-VL-30B-A3B-Think~\cite{bai2025qwen3vl}      & 60.7 &	23.9&	68.2&	31.3&	22.2&	17.1&	8.5&	5.5&	47.9&	22.0 \\
Qwen3-VL-235B-A22B-Think~\cite{bai2025qwen3vl}      & 74.7&	37.7&	84.4&	45.0&	31.4&	24.9&	14.1&	10.3&	60.9&	33.5 \\
InternVL3.5-8B-Think~\cite{wang2025internvl35} & 53.7&	18.0&	63.8&	21.6&	18.4&	13.7&	9.6&	6.5&	43.8&	16.5 \\
InternVL3.5-14B-Think~\cite{wang2025internvl35} & 55.9&	18.7&	67.1&	24.7&	19.4&	14.7&	12.3&	7.5&	46.4&	18.1 \\
InternVL3.5-30B-A3B-Think~\cite{wang2025internvl35}  & 54.4 &	18.0&	67.8&	26.1&	25.6&	17.1&	11.9&	7.6&	47.0&	18.7 \\
InternVL3.5-38B-Think~\cite{wang2025internvl35} & 58.8 &	21.4&	71.2&	24.8&	18.3&	11.6&	16.4&	10.5&	49.1&	19.1 \\
InternVL3.5-241B-A28B-Think~\cite{wang2025internvl35} & 64.9&	22.7&	76.4&	33.8&	20.5&	15.8&	11.8&	6.5&	52.5&	22.3 \\
Qwen3.5-9B-Think~\cite{qwen2026qwen35} & 67.1&	35.9&	77.6&	39.8&	39.1&	32.0&	16.9&	11.9&	56.0&	32.2 \\
Qwen3.5-27B-Think~\cite{qwen2026qwen35} & 74.0&	40.1&	82.2&	46.3&	42.1&	34.0&	24.7&	20.0&	62.5&	37.5 \\
Qwen3.5-35B-A3B-Think~\cite{qwen2026qwen35} & 72.4&	37.9&	85.2&	49.1&	38.2&	31.5&	23.3&	19.8&	62.8&	37.2 \\
Qwen3.5-122B-A10B-Think~\cite{qwen2026qwen35} & 77.0&	45.6&	84.2&	50.1&	46.5&	39.4&	23.9&	19.7&	65.2&	41.7 \\
Qwen3.5-397B-A17B-Think~\cite{qwen2026qwen35} & \textbf{79.4} &	\textbf{48.5} &	\textbf{86.0} &	\textbf{52.8} &	\textbf{52.1} &	\textbf{44.9} &	\textbf{33.0} &	\textbf{28.2} &	\textbf{69.6}&	\textbf{46.1} \\

\bottomrule
\end{tabular}
}
\end{table*}

\paragraph{Overall results.}

Table~\ref{tab:main_results} reports the main results on Video-IFBench. Gemini-3-Pro achieves the strongest overall performance, with 76.5 TCSR and 54.5 TISR. Among open-source models, Qwen3.5-397B-A17B-Think performs best, reaching 69.6 TCSR and 46.1 TISR and leaving a gap of 6.9 TCSR points and 8.4 TISR points to Gemini-3-Pro, respectively. Across all models, TISR is consistently lower than TCSR, indicating that models often satisfy some constraints but fail to meet all requirements in an instruction.

The results also reveal clear differences across instruction types. Many models achieve higher scores on Multi instructions than on Single instructions, indicating that the main bottleneck is not necessarily handling multiple tasks in one request, but rather following numerous and complex constraints. The challenge becomes more pronounced for Selection and Nested instructions, where models must first select the correct branch or path based on video content before executing the corresponding instruction. Nested instructions are particularly difficult: even Gemini-3-Pro obtains only 53.7 TCSR and 46.0 TISR, while the best open-source model reaches 33.0 TCSR and 28.2 TISR. And within the same model family, larger models generally exhibit better performance, as shown in Fig.~\ref{fig:family_scaling}.

\begin{figure}[t]
    \centering
    \makebox[\linewidth][c]{%
        \includegraphics[width=1.0\linewidth]{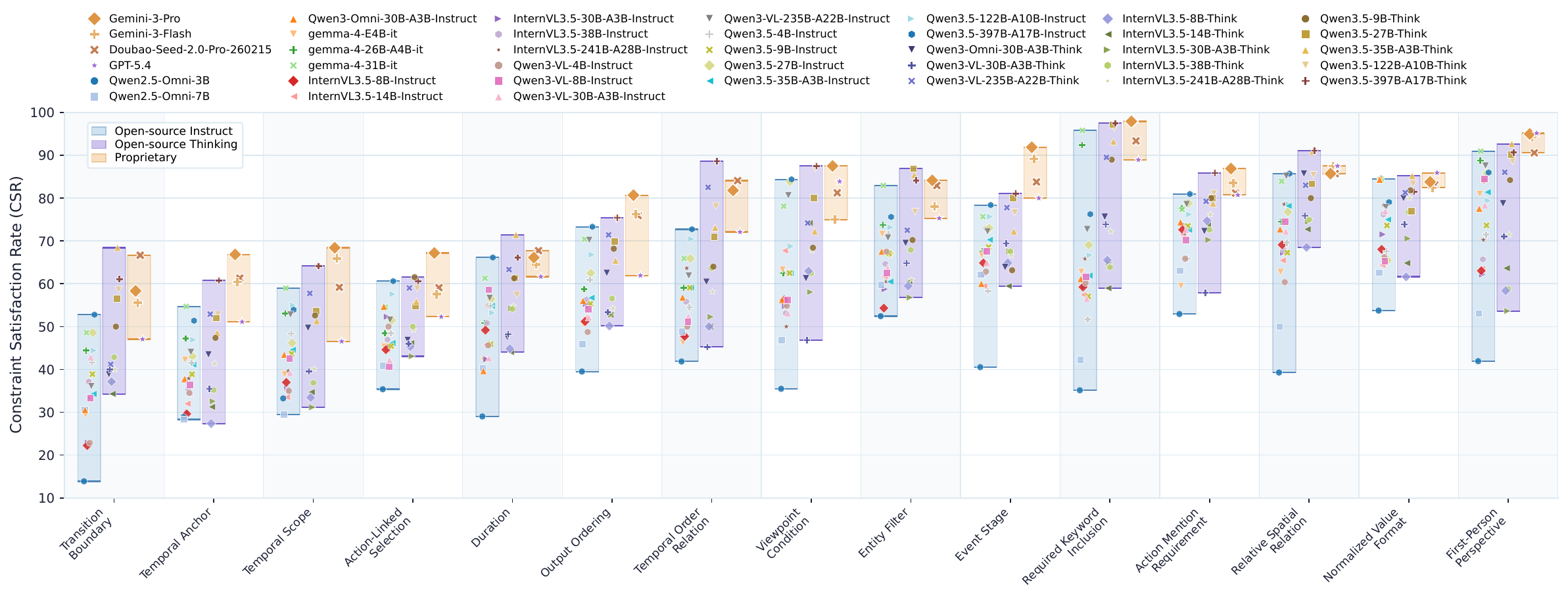}
    }
    \caption{Performance distribution on $15$ most challenging constraints. Colored bars show the performance range of different model groups, and points denote individual models.}
    \label{fig:constraint_model_spread}
    \vspace{-1em}
\end{figure}
\subsection{Further Analysis and Experiments}

\subsubsection{Constraint-level Performance Analysis}

We analyze constraint-level performance using Single and Selection instructions. Specifically, we compute the average CSR of all models for each constraint and report the 15 most challenging constraints in Fig.~\ref{fig:constraint_model_spread}. The results show that constraints related to temporal grounding and condition-dependent content selection, such as Transition Boundary, Temporal Anchor, Temporal Scope, and Action-Relative Selection, consistently achieve low scores across most models. This indicates that current models struggle to align their responses with fine-grained video evidence, especially when an instruction requires locating events or selecting content based on specific visual or audio conditions.

Fig.~\ref{fig:constraint_model_spread} also shows a clear separation among model groups. Proprietary models generally achieve higher scores on most constraints, while open-source models exhibit larger variation and lower performance. Among open-source models, thinking variants improve the upper bounds on most constraints. However, this improvement is not consistent across model families, a further analysis is provided in Appendix~\ref{appendix:effect_of_thinking_mode}.

\begin{figure*}[b]
    \centering
    \begin{subfigure}[t]{0.48\textwidth}
        \centering
        \includegraphics[width=\linewidth]{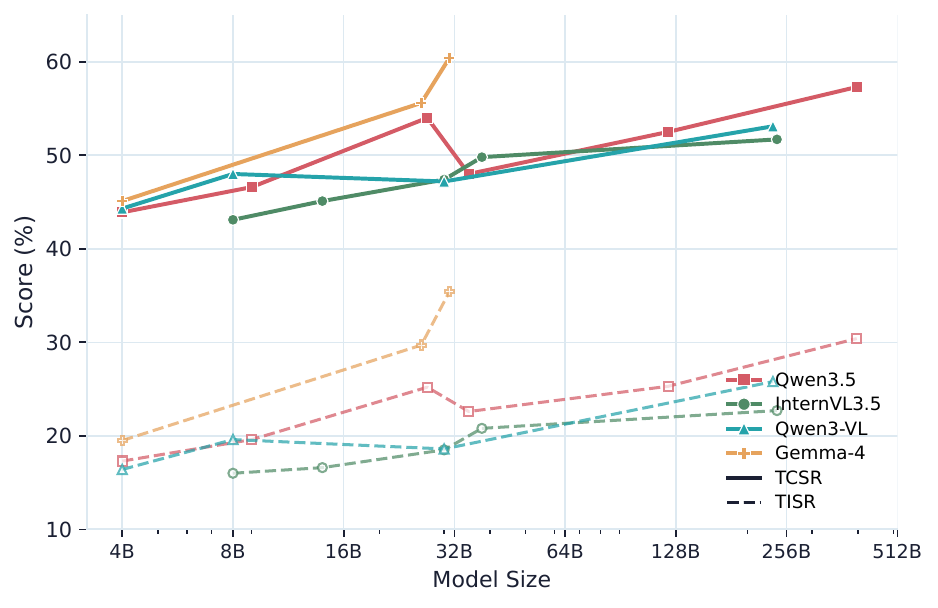}
        \caption{Scaling across model families.}
        \label{fig:family_scaling}
    \end{subfigure}
    \hfill
    \begin{subfigure}[t]{0.48\textwidth}
        \centering
        \includegraphics[width=\linewidth]{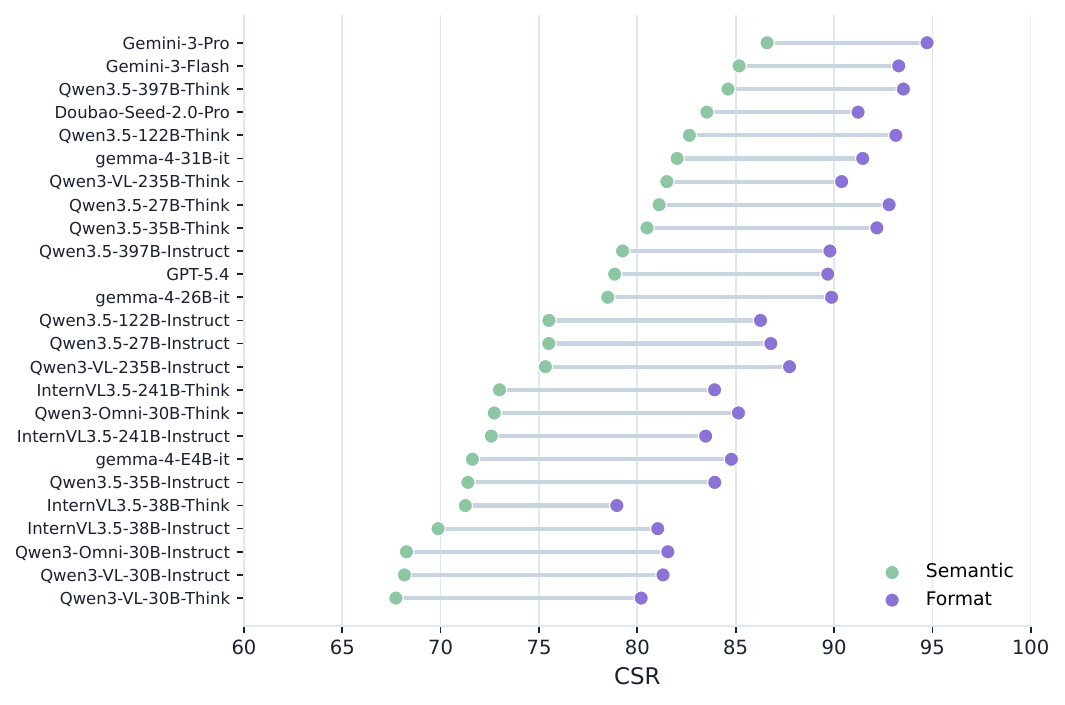}
        \caption{Semantic vs. format constraint performance.}
        \label{fig:semantic_format}
    \end{subfigure}

    \caption{Model scaling and semantic-format constraint performance.}
    \label{fig:scaling_semantic_format}
\end{figure*}

We further compare performance on semantic and format constraints in Fig.~\ref{fig:semantic_format}. Existing models perform substantially worse on semantic constraints than on format constraints, indicating that semantic constraints are more challenging because they require precise video understanding rather than merely formatting responses.

\subsubsection{Task and Constraint Scaling Analysis}
We analyze how performance changes with task and constraint complexity under controlled settings. For task scaling, we use Single and Multi instructions to avoid the confounding effect of conditional branching. For constraint scaling, we use Single instructions to minimize the influence of other factors. We report representative models covering proprietary models, open-source thinking and instruct models, omni-modal models, and different model scales.

\paragraph{Task scaling.}
Table~\ref{tab:task_scaling_placeholder} reports task pass rates as the number of tasks increases. Capable models remain relatively stable across task counts, suggesting that task count itself is not the main bottleneck. This observation is consistent with the findings in Section~\ref{sec:main_results}, which suggest that the primary challenge lies in adhering to constraints rather than handling multiple tasks.

\begin{table*}[t]
\centering
\scriptsize
\caption{Scaling analysis of task and constraint complexity. The left table reports task pass rate across task-count ranges, and the right table reports TISR across constraint-count ranges.}
\label{tab:scaling_analysis}

\begin{subtable}[t]{0.49\textwidth}
\centering
\caption{Task scaling on Single and Multi instructions.}
\label{tab:task_scaling_placeholder}
\resizebox{\linewidth}{!}{%
\begin{tabular}{lccc}
\toprule
Model & 1--4 & 5--8 & 9+ \\
\midrule
Gemini-3-Pro & 95.6 & 94.9 & 96.3 \\
Gemini-3-Flash & 93.9 & 95.7 & 94.3 \\
Qwen3.5-397B-A17B-Think & 95.5 & 92.6 & 95.8 \\
Gemma-4-31B-it & 90.5 & 92.5 & 91.8 \\
Qwen3-Omni-30B-A3B-Instruct & 84.5 & 81.0 & 76.9 \\
Gemma-4-E4B-it & 81.0 & 79.1 & 84.1 \\
\bottomrule
\end{tabular}
}
\end{subtable}
\hfill
\begin{subtable}[t]{0.49\textwidth}
\centering
\caption{Constraint scaling on Single instructions.}
\label{tab:constraint_scaling_placeholder}
\resizebox{\linewidth}{!}{%
\begin{tabular}{lccc}
\toprule
Model & 1--4 & 5--8 & 9+ \\
\midrule
Gemini-3-Pro & 62.7 & 51.8 & 46.9 \\
Gemini-3-Flash & 59.3 & 47.5 & 46.9 \\
Qwen3.5-397B-A17B-Think & 58.2 & 43.9 & 42.5 \\
Gemma-4-31B-it & 53.7 & 39.6 & 29.2 \\
Qwen3-Omni-30B-A3B-Instruct & 40.1 & 11.5 & 3.5 \\
Gemma-4-E4B-it & 38.4 & 12.9 & 7.1 \\
\bottomrule
\end{tabular}
}
\end{subtable}
\end{table*}

\paragraph{Constraint scaling.}
Table~\ref{tab:constraint_scaling_placeholder} shows that TISR drops substantially as the number of constraints increases. Gemini-3-Pro decreases from 62.7\% with 1--4 constraints to 46.9\% with 9+ constraints, while Qwen3.5-397B-A17B-Think drops from 58.2\% to 42.5\%. The decline is much sharper for weaker open-source models, with Qwen3-Omni-30B-A3B-Instruct falling from 40.1\% to 3.5\% and Gemma-4-E4B-it from 38.4\% to 7.1\%. These results show that current models struggle when many constraints are imposed within a Single instruction.


\subsubsection{Depth Sensitivity in Nested Instructions}

\begin{wrapfigure}{r}{0.62\linewidth}
    \centering
    \vspace{-0.8em}
    \includegraphics[width=1.0\linewidth]{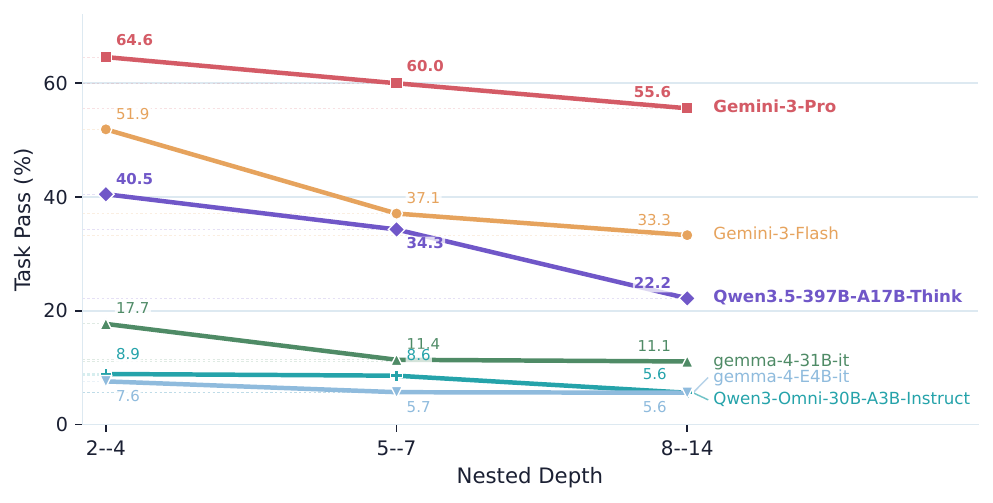}
    \caption{Depth sensitivity on Nested instructions.}
    \label{fig:nested_depth_sensitivity}
    \vspace{-1.0em}
\end{wrapfigure}

We analyze depth sensitivity on Nested instructions using the same representative models. As shown in Fig.~\ref{fig:nested_depth_sensitivity}, task pass rate generally decreases as the maximum tree depth increases. The drop is clear for strong models: Gemini-3-Pro decreases from 64.6\% at depth 2--4 to 55.6\% at depth 8--14, while Qwen3.5-397B-A17B-Think drops from 40.5\% to 22.2\%. Weaker open-source models remain low across all depth ranges, suggesting that they already struggle with shallow nested structures. These results show that deeper instruction trees make it harder for models to recognize the correct conditional path.


\subsubsection{Branch Number and Position in Selection Instructions}

\begin{table*}[t]
\centering
\scriptsize
\scriptsize
\caption{Task pass rate on Selection instructions using the 3-condition-plus-else subset.}
\label{tab:selection_ablation}
\begin{subtable}[t]{0.49\textwidth}
\centering
\caption{Branch count.}
\resizebox{\linewidth}{!}{%
\begin{tabular}{lccc}
\toprule
Model & 2 & 3 & 4 \\
\midrule
Gemini-3-Flash & 66.7 & 58.0 & 56.5 \\
Qwen3.5-397B-A17B-Think & 68.1 & 59.4 & 59.4 \\
Qwen3.5-397B-A17B-Instruct & 58.0 & 50.7 & 39.1 \\
Qwen3.5-9B-Instruct & 37.7 & 24.6 & 27.5 \\
Gemma-4-31B-it & 50.7 & 46.4 & 36.2 \\
Gemma-4-E4B-it & 43.5 & 27.5 & 19.1 \\
Qwen3-Omni-30B-A3B-Instruct & 43.9 & 30.4 & 17.4 \\
\bottomrule
\end{tabular}
}
\end{subtable}
\hfill
\begin{subtable}[t]{0.49\textwidth}
\centering
\caption{Correct-branch position.}
\resizebox{\linewidth}{!}{%
\begin{tabular}{lccc}
\toprule
Model & 1 & 2 & 3 \\
\midrule
Gemini-3-Flash & 66.7 & 66.7 & 54.9 \\
Qwen3.5-397B-A17B-Think & 76.5 & 58.8 & 52.9 \\
Qwen3.5-397B-A17B-Instruct & 70.6 & 35.3 & 29.4 \\
Qwen3.5-9B-Instruct & 45.1 & 21.6 & 15.7 \\
Gemma-4-31B-it & 33.3 & 43.1 & 39.2 \\
Gemma-4-E4B-it & 33.3 & 8.3 & 5.9 \\
Qwen3-Omni-30B-A3B-Instruct & 23.5 & 12.0 & 18.0 \\
\bottomrule
\end{tabular}
}
\end{subtable}
\end{table*}

We conduct controlled ablation studies on the 3-condition-plus-else subset to study how branch count and correct-branch position affect performance on Selection instructions. For branch count analysis, we keep the correct branch unchanged and progressively add one or two false branches together with the else branch, resulting in 2, 3, and 4-branch variants. For position analysis, we keep the same set of condition branches but move the correct branch to different positions. 

As shown in Table~\ref{tab:selection_ablation}, increasing the number of candidate branches generally lowers task pass rate. For example, Gemini-3-Flash drops from 66.7\% with two branches to 56.5\% with four branches, and Qwen3.5-397B-A17B-Instruct drops from 58.0\% to 39.1\%. This suggests that models become less reliable as they must compare more video-grounded conditions.

We find that the correct-branch position has a clear impact. Most models show a preference for the first branch, with performance decreasing when the correct branch appears later. For example, Qwen3.5-397B-A17B-Think drops from 76.5\% at position 1 to 52.9\% at position 3, while its Instruct counterpart drops from 70.6\% to 29.4\%. These results indicate that current models do not evaluate conditional branches equally and may rely on positional biases.

\section{Conclusion}
We introduce Video-IFBench, a benchmark for evaluating instruction following in video understanding, with a hybrid protocol combining LLM-as-Judge and programmatic verification. Experiments show that current MLLMs struggle with constraint-rich and conditional video instructions. We hope Video-IFBench will facilitate future research on video MLLMs that can faithfully follow complex, constraint-rich instructions in practical applications.

\bibliographystyle{plain}
\bibliography{reference}

\newpage
\appendix


\section{Additional Experimental Results}
\subsection{Main results using Qwen3.5-35B-A3B as Judge Model.}
\label{appendix:qwen3_5_35b_judge}
To examine the effect of judge model choice, we re-evaluate all model responses using Qwen3.5-35B-A3B as the judge. Table~\ref{tab:qwen35_judge_main_results} reports the resulting scores. The values in parentheses in the Overall columns denote the difference from the main results in Table~\ref{tab:main_results}. Using a smaller judge generally increases the absolute scores, but the relative trends remain largely consistent with the main results, supporting the robustness of our evaluation protocol.

\begin{table*}[t]
\centering
\scriptsize
\setlength{\tabcolsep}{4pt}
\renewcommand{\arraystretch}{1.0}
\caption{Main results evaluated with Qwen3.5-35B-A3B as the judge. Best results within each model group are in bold. Parenthesized values in the Overall columns denote changes from the main results.}
\label{tab:qwen35_judge_main_results}
\resizebox{\textwidth}{!}{
\begin{tabular}{lcccccccccc}
\toprule
\multirow{2}{*}{Model}
& \multicolumn{2}{c}{Single}
& \multicolumn{2}{c}{Multi}
& \multicolumn{2}{c}{Selection}
& \multicolumn{2}{c}{Nested}
& \multicolumn{2}{c}{Overall} \\
\cmidrule(lr){2-3} \cmidrule(lr){4-5} \cmidrule(lr){6-7} \cmidrule(lr){8-9} \cmidrule(lr){10-11}
& TCSR & TISR & TCSR & TISR & TCSR & TISR & TCSR & TISR & TCSR & TISR \\
\midrule
\rowcolor{groupgray}
\multicolumn{11}{c}{\textbf{Proprietary Models}} \\
\midrule
Gemini-3-Pro~\cite{google2025gemini3} & \textbf{86.6} & \textbf{64.1} & 87.7 & 53.3 & \textbf{73.1} & \textbf{63.0} & \textbf{57.8} & \textbf{51.5} & \textbf{80.1 (+3.6)} & \textbf{58.7 (+4.2)} \\
Gemini-3-Flash~\cite{google2025gemini3} & 83.0 & 55.8 & \textbf{87.8} & \textbf{53.4} & 67.1 & 57.3 & 43.8 & 35.3 & 75.7 (+3.5) & 52.2 (+2.7) \\
Doubao-Seed-2.0-Pro-260215~\cite{bytedanceseed2026seed20} & 84.1 & 54.3 & 85.9 & 48.7 & 47.7 & 41.2 & 20.8 & 17.8 & 68.6 (+2.9) & 44.7 (+3.9) \\
GPT-5.4~\cite{singh2025gpt5} & 78.0 & 42.5 & 80.1 & 40.3 & 24.7 & 20.4 & 13.5 & 9.9 & 59.6 (+2.2) & 33.0 (+3.0) \\
\midrule
\rowcolor{groupgray}
\multicolumn{11}{c}{\textbf{Open-source Models (Instruct)}} \\
\midrule
Qwen2.5-Omni-3B~\cite{xu2025qwen25omni} & 40.9 & 13.6 & 32.0 & 8.3 & 10.0 & 6.7 & 7.8 & 4.8 & 28.0 (+2.2) & 9.5 (+0.9) \\
Qwen2.5-Omni-7B~\cite{xu2025qwen25omni} & 53.1 & 19.8 & 48.5 & 13.9 & 16.7 & 10.6 & 12.0 & 7.7 & 39.1 (+3.1) & 14.6 (+1.1) \\
Qwen3-Omni-30B-A3B-Instruct~\cite{xu2025qwen3omni} & 67.6 & 27.8 & 64.8 & 25.9 & 20.7 & 15.9 & 7.9 & 5.5 & 49.0 (+4.7) & 21.6 (+3.6) \\
Gemma-4-E4B-it~\cite{google2026gemma4} & 65.0 & 28.8 & 65.8 & 27.7 & 14.2 & 9.6 & 8.0 & 5.5 & 48.2 (+3.1) & 21.8 (+2.3) \\
Gemma-4-26B-A4B-it~\cite{google2026gemma4} & 77.6 & 44.7 & 78.0 & 39.4 & 25.6 & 21.1 & 10.6 & 7.4 & 58.4 (+2.8) & 33.3 (+3.6) \\
Gemma-4-31B-it~\cite{google2026gemma4} & \textbf{81.7} & \textbf{50.1} & 80.9 & \textbf{42.1} & \textbf{34.8} & \textbf{31.1} & 17.3 & \textbf{13.4} & \textbf{63.5 (+3.1)} & \textbf{38.8 (+3.4)} \\
InternVL3.5-8B-Instruct~\cite{wang2025internvl35} & 63.9 & 24.1 & 63.1 & 23.0 & 19.8 & 13.6 & 9.9 & 5.4 & 47.7 (+4.6) & 19.1 (+3.1) \\
InternVL3.5-14B-Instruct~\cite{wang2025internvl35} & 65.1 & 24.6 & 68.2 & 25.5 & 19.7 & 12.5 & 10.3 & 7.5 & 49.8 (+4.7) & 20.2 (+3.6) \\
InternVL3.5-30B-A3B-Instruct~\cite{wang2025internvl35} & 66.2 & 24.7 & 65.7 & 23.1 & 27.6 & 20.1 & 18.1 & 11.7 & 52.0 (+4.6) & 21.5 (+3.0) \\
InternVL3.5-38B-Instruct~\cite{wang2025internvl35} & 68.9 & 28.2 & 71.9 & 24.9 & 22.3 & 15.9 & \textbf{18.6} & 12.4 & 54.0 (+4.2) & 22.6 (+1.8) \\
InternVL3.5-241B-A28B-Instruct~\cite{wang2025internvl35} & 71.7 & 30.6 & 74.3 & 30.2 & 19.6 & 15.0 & 15.6 & 10.9 & 54.8 (+3.1) & 24.8 (+2.1) \\
Qwen3-VL-4B-Instruct~\cite{bai2025qwen3vl} & 64.2 & 23.1 & 62.5 & 21.9 & 18.2 & 13.9 & 11.1 & 5.9 & 47.4 (+3.1) & 18.5 (+2.1) \\
Qwen3-VL-8B-Instruct~\cite{bai2025qwen3vl} & 67.7 & 27.4 & 69.1 & 26.5 & 22.1 & 16.8 & 15.5 & 10.4 & 51.8 (+3.8) & 22.6 (+3.0) \\
Qwen3-VL-30B-A3B-Instruct~\cite{bai2025qwen3vl} & 67.6 & 25.1 & 69.5 & 25.9 & 18.0 & 11.9 & 14.3 & 10.0 & 51.1 (+3.9) & 20.6 (+2.0) \\
Qwen3-VL-235B-A22B-Instruct~\cite{bai2025qwen3vl} & 75.0 & 36.3 & 79.8 & 37.0 & 20.6 & 16.7 & 9.6 & 6.4 & 57.0 (+3.9) & 28.5 (+2.7) \\
Qwen3.5-4B-Instruct~\cite{qwen2026qwen35} & 63.8 & 25.4 & 64.4 & 23.6 & 18.9 & 13.3 & 10.0 & 5.5 & 47.9 (+4.0) & 19.7 (+2.4) \\
Qwen3.5-9B-Instruct~\cite{qwen2026qwen35} & 65.5 & 26.5 & 67.4 & 28.3 & 18.4 & 15.0 & 10.1 & 5.5 & 49.2 (+2.6) & 21.8 (+2.2) \\
Qwen3.5-27B-Instruct~\cite{qwen2026qwen35} & 75.7 & 35.9 & 76.8 & 35.8 & 26.2 & 20.2 & 13.9 & 8.5 & 57.9 (+3.9) & 28.9 (+3.7) \\
Qwen3.5-35B-A3B-Instruct~\cite{qwen2026qwen35} & 67.9 & 29.9 & 72.7 & 31.1 & 20.1 & 14.5 & 10.9 & 7.0 & 52.1 (+4.1) & 24.0 (+1.4) \\
Qwen3.5-122B-A10B-Instruct~\cite{qwen2026qwen35} & 70.9 & 33.1 & 76.9 & 35.4 & 23.3 & 16.2 & 13.6 & 8.5 & 55.7 (+3.2) & 27.1 (+1.8) \\
Qwen3.5-397B-A17B-Instruct~\cite{qwen2026qwen35} & 79.9 & 43.9 & \textbf{80.9} & 38.3 & 28.3 & 21.6 & 14.3 & 9.0 & 61.3 (+4.0) & 33.1 (+2.7) \\
\midrule
\rowcolor{groupgray}
\multicolumn{11}{c}{\textbf{Open-source Models (Thinking)}} \\
\midrule
Qwen3-Omni-30B-A3B-Think~\cite{xu2025qwen3omni} & 75.0 & 36.0 & 76.2 & 35.4 & 30.3 & 23.0 & 11.7 & 7.0 & 57.8 (+4.7) & 29.1 (+2.9) \\
Qwen3-VL-30B-A3B-Think~\cite{bai2025qwen3vl} & 65.4 & 29.3 & 69.2 & 30.2 & 26.0 & 21.2 & 10.8 & 6.7 & 51.1 (+3.2) & 24.6 (+2.6) \\
Qwen3-VL-235B-A22B-Think~\cite{bai2025qwen3vl} & 81.4 & 46.6 & 81.7 & 43.2 & 33.9 & 27.6 & 16.8 & 13.3 & 63.3 (+2.4) & 37.2 (+3.7) \\
InternVL3.5-8B-Think~\cite{wang2025internvl35} & 62.1 & 21.1 & 61.4 & 24.7 & 21.6 & 14.5 & 12.8 & 9.0 & 47.1 (+3.3) & 19.1 (+2.6) \\
InternVL3.5-14B-Think~\cite{wang2025internvl35} & 64.9 & 24.6 & 66.4 & 22.3 & 24.5 & 16.0 & 17.5 & 10.1 & 51.0 (+4.6) & 20.2 (+2.1) \\
InternVL3.5-30B-A3B-Think~\cite{wang2025internvl35} & 62.4 & 21.0 & 64.4 & 24.3 & 27.1 & 19.3 & 15.7 & 10.6 & 49.6 (+2.6) & 20.1 (+1.4) \\
InternVL3.5-38B-Think~\cite{wang2025internvl35} & 67.1 & 24.1 & 68.4 & 23.7 & 22.6 & 15.5 & 19.4 & 11.6 & 52.5 (+3.4) & 20.6 (+1.5) \\
InternVL3.5-241B-A28B-Think~\cite{wang2025internvl35} & 72.9 & 30.9 & 75.2 & 33.0 & 23.0 & 17.5 & 13.8 & 8.5 & 55.9 (+3.4) & 25.8 (+3.5) \\
Qwen3.5-9B-Think~\cite{qwen2026qwen35} & 77.8 & 40.6 & 78.1 & 37.6 & 43.1 & 37.8 & 17.8 & 13.5 & 61.3 (+5.3) & 34.9 (+2.7) \\
Qwen3.5-27B-Think~\cite{qwen2026qwen35} & 83.3 & 47.8 & 84.4 & 46.2 & 44.6 & 37.2 & 30.3 & 26.5 & 67.9 (+5.4) & 42.0 (+4.5) \\
Qwen3.5-35B-A3B-Think~\cite{qwen2026qwen35} & 81.1 & 45.8 & 82.9 & 48.1 & 39.5 & 30.9 & 26.4 & 24.3 & 66.4 (+3.6) & 40.7 (+3.5) \\
Qwen3.5-122B-A10B-Think~\cite{qwen2026qwen35} & 84.1 & 55.9 & 85.0 & \textbf{50.5} & 49.3 & 43.8 & 26.6 & 23.6 & 68.9 (+3.7) & 47.0 (+5.3) \\
Qwen3.5-397B-A17B-Think~\cite{qwen2026qwen35} & \textbf{86.2} & \textbf{59.6} & \textbf{85.8} & 48.3 & \textbf{55.7} & \textbf{47.8} & \textbf{35.5} & \textbf{31.2} & \textbf{73.0 (+3.4)} & \textbf{49.8 (+3.7)} \\

\bottomrule
\end{tabular}
}
\end{table*}

\subsection{Effect of Thinking Mode on Video-IFBench.}
\label{appendix:effect_of_thinking_mode}
\begin{figure}[t]
    \centering
    \includegraphics[width=\linewidth]{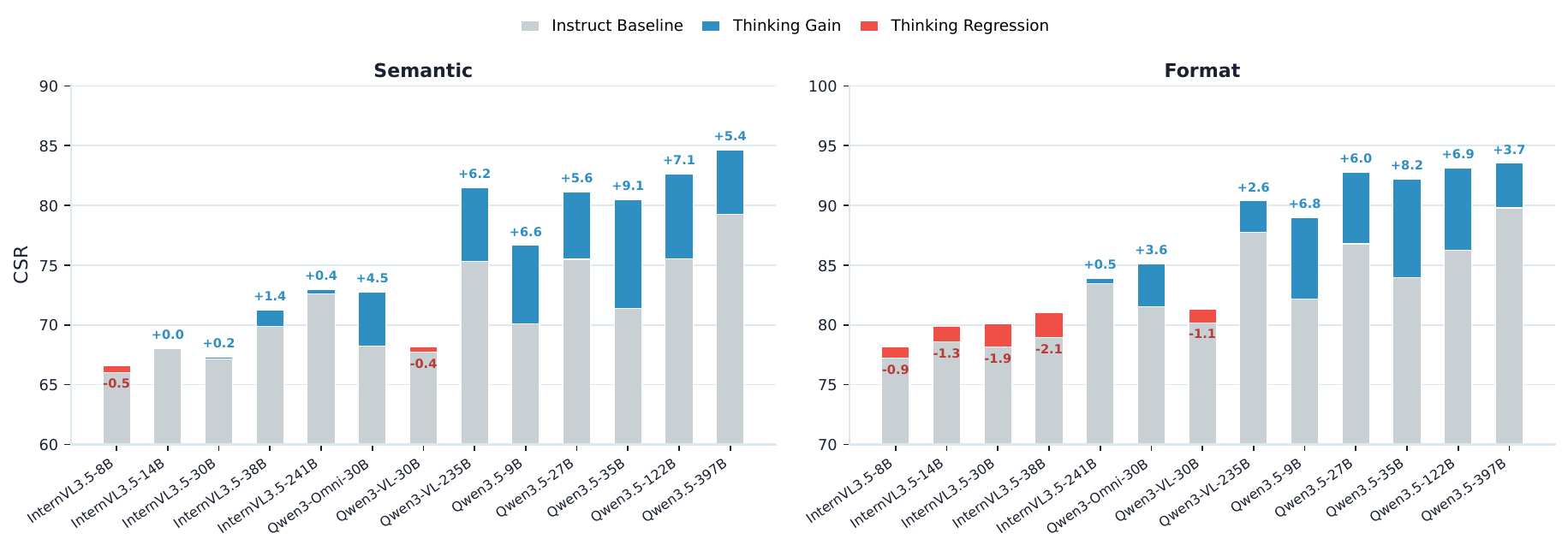}
    \caption{Effect of thinking mode on semantic and format constraints.}
    \label{fig:thinking_gain}
\end{figure}

As shown in Fig.~\ref{fig:thinking_gain}, thinking mode brings clear gains for Qwen3 and Qwen3.5 models, but yields little improvement or even performance drops for InternVL3.5. We hypothesize that the model-dependent effect may be related to how instruction-following rewards are designed during reinforcement learning. Moreover, the gains are generally larger on semantic constraints than on format constraints, suggesting that thinking-style inference is more helpful when constraint following requires visual or audio semantic understanding.

\section{Task Taxonomy}
\label{appendix:task_taxonomy}
\begingroup
\scriptsize
\setlength{\tabcolsep}{3pt}
\renewcommand{\arraystretch}{1.03}
\begin{longtable}{>{\raggedright\arraybackslash}p{0.15\linewidth} >{\raggedright\arraybackslash}p{0.18\linewidth} >{\raggedright\arraybackslash}p{0.40\linewidth} >{\raggedright\arraybackslash}p{0.21\linewidth}}
\caption{Video understanding task taxonomy in Video-IFBench.}\label{tab:task_taxonomy}\\
\toprule
Category & Task Type & Description & Example Question \\
\hline
\endfirsthead
\toprule
Category & Task Type & Description & Example Question \\
\hline
\endhead
Perception \& Recognition & Entity Recognition & To describe specific objects, individuals, animals, or elements present within the video frames. & What type of vehicle is parked in front of the house at 0:15? \\
\hline
Perception \& Recognition & Action Recognition & To classify the specific movements, gestures, or activities being performed by entities in the video. & What is the man in the blue shirt doing when the black car stops? \\
\hline
Perception \& Recognition & Attribute Recognition & To describe specific properties or characteristics of an entity, such as its color, size, shape, or material. & What color is the backpack the child is carrying? \\
\hline
Perception \& Recognition & Scene Description & To recognize and describe the global environment, setting, or background where the video takes place. & Count the number of scenes appeared in the video and describe each of them in detail. \\
\hline
Perception \& Recognition & Text Recognition (OCR) & To detect, extract, and read visible text within the video, such as street signs, subtitles, documents, or labels. & What does the neon sign above the shop door say? \\
\hline
Perception \& Recognition & Dialogue Recognition & To recognize and describe spoken content, speaker turns, or dialogue exchanged in the video. & What are the exact words the news anchor says right after the loud crash is heard? \\
\hline
Perception \& Recognition & Counting & To determine how many times a target speech pattern occurs or how many target objects appear throughout the video. & How many times does the host say 'thank you' in the video? \\
\hline
Perception \& Recognition & Emotion Recognition & To infer the emotional state, mood, or facial expressions of the people or characters in the video. & Does the woman look happy or frustrated when she opens the cardboard box? \\
\hline
Perception \& Recognition & Event Recognition & To categorize the overarching macroscopic event or social activity happening in the video clip. & Describe, in chronological order, the events experienced by the man wearing red in the video. \\
\hline
Temporal Understanding & Event Grounding & To pinpoint the exact start and end timestamps of a specific event or action, including all instances of its occurrence. & At what exact timestamps does the dog bark during the video? \\
\hline
Temporal Understanding & Occurrence Localization & To locate the precise time interval when an entity, action, or state appears for the first, last, or N-th time. & When does the blue car appear for the third time in the footage? \\
\hline
Temporal Understanding & Temporal Ordering & To determine the chronological sequence of two or more distinct actions or events. & Did the character put on his hat before or after he picked up the umbrella? \\
\hline
Temporal Understanding & Temporal Adjacent Retrieval & To retrieve the event or action that occurs immediately preceding or following a specified anchor event. & What happens immediately after the glass shatters on the floor? \\
\hline
Temporal Understanding & Local Temporal Understanding & To sequentially narrate the list of actions or events that take place within a strictly defined time window. & List all the actions performed by the chef between 1:00 and 1:30 in chronological order. \\
\hline
Temporal Understanding & Temporal Sorting & To logically reorder a randomized or shuffled list of video events into their correct chronological sequence. & Given the events 'eating dinner', 'cooking food', and 'washing dishes', order them as they actually occurred in the video. \\
\hline
Temporal Understanding & Duration Estimation & To calculate or estimate the length of time a specific action, event, or state persists. & How many seconds does the traffic light remain red? \\
\hline
Temporal Understanding & Audio Change & To identify and describe how spoken audio evolves over time, such as changes in active speaker, speaking state, turn-taking, or stretches of silence. & Describe how the active speaker changes from the beginning of the video to the end. \\
\hline
Temporal Understanding & Audio-Visual Temporal Alignment & To determine and describe the temporal correspondence between spoken audio cues and visual events, identifying which utterances or speaker turns align with which visible actions, moments, or scene transitions. & Describe which visible actions coincide with the moments when the host starts speaking. \\
\hline
Spatial Understanding & Relative Direction Judgment & To determine the directional position of an entity relative to the camera's viewpoint or another reference entity. & Describe the clothing of all people to the left of the entrance. \\
\hline
Spatial Understanding & Relative Distance Comparison & To compare and determine which entity is closer to or further from a given reference point. & Order the oak tree, park bench, motorcycle, and statue from farthest to nearest relative to the black trash can. \\
\hline
Spatial Understanding & Spatial Relationship Judgment & To classify the spatial relations between entities, such as inside, on, under, beside, or attached to. & Describe the cat’s spatial relationship to the table, chair, rug, and fireplace while it is sleeping in the video. \\
\hline
Spatial Understanding & Perspective-based Spatial Judgment & To reinterpret spatial relationships from a hypothetical or shifted orientation (e.g., 'standing at X facing Y'). & If you are standing at the doorway facing the window, describe all the objects on your left-hand side and their relative positions. \\
\hline
Relationship \& Interaction & Entity Interaction Recognition & To describe dynamic interactions between multiple entities, such as holding, passing, attacking, talking, or hugging. & How do the two players interact immediately after the goal is scored? \\
\hline
Relationship \& Interaction & Action-Entity Binding & To accurately associate actions with their specific agents and targets to determine who did what to whom. & Describe the attributes of the player who passes the basketball to the player wearing a yellow jersey. \\
\hline
Relationship \& Interaction & Speech Source Recognition & To visually localize the source of spoken audio, such as identifying which person in a crowd is speaking. & Which person on screen is speaking when the announcement is heard? \\
\hline
Logical Reasoning & Causal Reasoning & To deduce the cause-and-effect relationship between different events, actions, or states in the video. & Why did the fire alarm go off in the hallway? \\
\hline
Logical Reasoning & Counterfactual Reasoning & To imagine and reason about alternative outcomes if a specific past event had not occurred or happened differently. & If the man hadn't dropped his keys on the sidewalk, would he have missed the bus? \\
\hline
Logical Reasoning & Multi-hop Reasoning & To answer complex queries by synthesizing evidence across multiple timestamps, events, or logical steps. & Based on the ingredients the chef gathered at the start and the cooking method used in the middle, what specific dish will be plated at the end? \\
\hline
Logical Reasoning & Knowledge Transfer & To extract a rule, concept, or physical principle shown in the video and apply it to a new, unseen hypothetical scenario. & Based on the physics principle demonstrated with the pendulum in the video, what would happen if the string were twice as long? \\
\hline
Captioning & Segmented Captioning & To generate individual, coherent textual descriptions for pre-defined or uniformly segmented time intervals of the video. & Provide a descriptive caption for the events occurring between 0:00-0:30, 0:30-1:00, and 1:00-1:30. \\
\hline
Captioning & Dense Captioning & To generate highly detailed, fine-grained descriptions covering all visible actions, entities, and micro-events throughout the video. & Provide a dense, second-by-second narrative of every action happening in this street view camera footage. \\
\hline
Captioning & Video Summarization & To generate a concise, high-level overview that captures the core theme, main plot, or key highlights of the entire video. & Write a two-sentence summary of the main story and outcome shown in this short film. \\
\hline
\end{longtable}
\endgroup

\section{Constraint Taxonomy}
\label{appendix:constraint_taxonomy}
\begingroup
\scriptsize
\setlength{\tabcolsep}{3pt}
\renewcommand{\arraystretch}{1.03}
\begin{longtable}{>{\raggedright\arraybackslash}p{0.10\linewidth} >{\raggedright\arraybackslash}p{0.17\linewidth} >{\raggedright\arraybackslash}p{0.45\linewidth} >{\raggedright\arraybackslash}p{0.22\linewidth}}
\caption{Constraint taxonomy in Video-IFBench.}\label{tab:constraint_taxonomy}\\
\toprule
Group & Constraint Type & Description & Example \\
\hline
\endfirsthead
\toprule
Group & Constraint Type & Description & Example \\
\hline
\endhead
Semantic & Temporal Scope & Restrict the task to evidence, entities, events, or judgments within a specified absolute or segment-defined time window of the video, such as a timestamp range, beginning/middle/end portion, or last N seconds. & Use only the visual events that occur in the last 10 seconds of the video in your answer. \\
\hline
Semantic & Temporal Anchor & Constrain a target relative to one explicit reference event, action, state, or timestamp, such as before, after, during, or at the moment of that anchor. & Mention only the video content before the door opens. \\
\hline
Semantic & Temporal Order Relation & Require the response to determine or preserve a temporal ordering relation among two or more already-specified events or targets, such as whether one happens before, after, between, or immediately adjacent to another. & Indicate the temporal order of all mentioned visual events in your answer. \\
\hline
Semantic & Event Stage & Restrict the target to an intrinsic phase of a single continuous event or process, such as onset, peak, hold, completion, or recovery, rather than an arbitrary external time window. & Do not mention any video content after the man leaves the room in your answer. \\
\hline
Semantic & Duration Constraint & Require the target event, state, relation, or occurrence to satisfy a duration condition such as minimum length, maximum length, longest, shortest, brief, or sustained. & Mention only actions lasting more than 2 seconds. \\
\hline
Semantic & Temporal Granularity & Constrain the temporal unit used in the answer to a specified granularity, such as seconds only, minutes only, or shot indices only. & Use shot numbers for every mentioned visual segment in your answer; do not use other timestamps. \\
\hline
Semantic & Spatial Region Restriction & Restrict valid candidates to a designated region. & Do not mention any outdoor content in your answer. \\
\hline
Semantic & Relative Spatial Relation & Select or interpret the target based on its spatial relation to another object, person, landmark, or scene element. & For all stationary objects mentioned in your answer, describe their spatial relationships relative to the leftmost streetlight at the 3-second mark of the video. \\
\hline
Semantic & Viewpoint Condition & Require the judgment to be made only under a specified camera view, shot type, or visual perspective when interpretation depends on viewpoint. & From a bird’s-eye view, describe the positions of all objects mentioned in the answer. \\
\hline
Semantic & Category Filter & Restrict valid targets to a specified semantic category or predefined set of categories. & In the answer, only food-related entities may be mentioned; all other entities must be replaced with uppercase English letters starting from C. \\
\hline
Semantic & Entity Filter & Require the valid entity or speaker targets mentioned in the answer to satisfy a specified attribute, appearance property, acoustic trait, or state qualifier. & Mention only people wearing red jackets in the answer. \\
\hline
Semantic & Attribute Avoidance & Restrict the output from mentioning specific attributes, physical traits, acoustic traits, or sensitive characteristics of the entities. & Do not mention any person's gender or age in the answer. \\
\hline
Semantic & Attribute Mention Requirement & Require the answer to explicitly mention one or more specified attributes, appearance properties, acoustic traits, or state descriptors for each relevant entity or speaker mentioned in the response, rather than only selecting targets by those attributes. & For every person mentioned in the answer, include the color of their pants. \\
\hline
Semantic & Action-Relative Selection & Restrict valid targets to those participating in a specified action, interaction, role, or event involvement condition. & Only mention players who were involved in passing in the answer. \\
\hline
Semantic & Action Mention Requirement & Require the answer to explicitly mention one or more specified actions, interactions, or motion types performed by a designated entity, entity group, body part, or target set, optionally covering all valid instances rather than merely identifying participants. & The answer must include all actions performed by the woman in the video. \\
\hline
Semantic & Action Avoidance & Restrict the answer from mentioning specified actions, interactions, or motion types associated with a designated entity, entity group, body part, instrument, or target set, even if those actions are visible in the video. & Do not mention any actions performed by the woman's hands in the answer. \\
\hline
Semantic & Counting Scope & Constrain how many valid items, entities, events, or occurrences should be returned or counted, such as exactly k, at most k, all, or one. & Only two people may be explicitly mentioned in the answer; all others must be referred to using uppercase English letters. \\
\hline
Semantic & Transition Boundary & Focus on the boundary where presence, state, relation, or scene status changes, such as appears, disappears, turns on, becomes inside, or shifts environment. & The answer must include every moment when the man in blue enters or exits the frame. \\
\hline
Format & Output Ordering & Require outputs to follow a specified ordering rule such as chronological order, reading order, numeric order, order of appearance, or speaker-turn order. & All objects mentioned in the answer must be ordered from farthest to nearest relative to the man. \\
\hline
Format & Normalized Value Format & Constrain scalar, label, date, string, or top-k outputs to a specified normalized representation without imposing a full table, list, or JSON structure. & Return dates in [YYYY-MM-DD] format. \\
\hline
Format & Structured JSON Output & Require the answer to be a valid JSON object or JSON array, optionally with required top-level keys or a fixed schema skeleton. & Return the answer as a JSON object with keys reason and evidence\_segments. \\
\hline
Format & Table Output & Require the answer to be presented as a table with an explicit row-column structure, such as Markdown table or CSV-style table. & Summarize each stage in a two-column table. \\
\hline
Format & Ordered List Output & Require the answer to be organized as an ordered list or bullet list rather than a free-form paragraph. & List the key events as an ordered list. \\
\hline
Format & Closed-Set Option Output & Require the answer to be exactly one option from a provided closed set, optionally with no extra explanation. & Answer with exactly one of: yes, no, unknown. \\
\hline
Format & Response Language & Require the entire answer to be produced in a specified language or bilingual pair. & answer entirely in Spanish \\
\hline
Format & Response Length Constraint & Constrain the overall answer length by an exact value, a lower bound, an upper bound, or a closed range, typically measured in words, or in project-defined Chinese length units for Chinese responses. & Answer in 80 to 120 words. \\
\hline
Format & Paragraph Count Constraint & Require the answer to contain a specified number of paragraphs or paragraph-like blocks. & Write the explanation in exactly three paragraphs. \\
\hline
Format & List Item Count Constraint & Require a list-style answer to contain exactly or at most a specified number of bullet items or numbered items. & Give exactly five bullet points. \\
\hline
Format & Title Requirement & Require the answer to include a title with a specified style, scope, or placement. & add a short title before the summary \\
\hline
Format & Required Keyword Inclusion & Require the answer to include one or more specified words or phrases. & The summary must include the word touchdown exactly three times. \\
\hline
Format & Forbidden Keyword Exclusion & Require the answer to avoid a specified banned word set, phrase set, or lexical family. & Do not mention she, he, it, they, or we. \\
\hline
Format & First-Person Perspective & Require the answer to be phrased from a first-person point of view rather than an external narrator perspective. & Answer from the first-person perspective of the woman wearing red in the scene. \\
\hline
Format & Text Type Restriction & Restrict text, quoted speech, or symbol targets by semantic type, lexical pattern, script, style, or character class. & Any on-screen text quoted in the answer must be written entirely in uppercase. \\
\hline
Format & Entity Label Set Restriction & Restrict the entity, speaker, or category labels used within the answer to an explicitly given closed-set label inventory or a specified branch of a taxonomy, without requiring the entire answer to be a single closed-set option. & All entities mentioned in the answer may only be referred to using the following names: [Anna, John]. \\
\hline
Semantic & Source Restriction & Restrict the sources cited in the answer, such as particular speakers, visible text, on-screen graphics, documents, or other designated information sources. & The answer must cite only text shown on the scoreboard and must not include any spoken dialogue. \\
\hline
Semantic & Speaker-Speech Binding & Require every referenced speech item in the answer to be explicitly bound to a specific speaker. & Every speech item must be bound to a specific speaker. \\
\hline
Semantic & Speaker Localization Constraint & Require each referenced speaker in the answer to be localized using a coarse on-screen position or relative visual location. & For each speech item, indicate whether the speaker is on the left, center, or right side of the frame. \\
\hline
Semantic & Segment Localization & Require each referenced speech item or visual segment in the answer to be localized to a specific start timestamp and end timestamp in the video. & Every speech item must include its corresponding start timestamp and end timestamp. \\
\hline
Format & Temporal/Tense Constraint & Restrict the output to be generated using specific grammatical tenses or within a specific timeline context. & Use the past tense in the first sentence of the answer, and the past perfect tense in the rest. \\
\hline
\end{longtable}
\endgroup

\section{Example Visualization}
\label{appendix:example_vis}
\begin{figure}[!htbp]
    \centering
    \makebox[\linewidth][c]{%
        \includegraphics[width=1.0\linewidth]{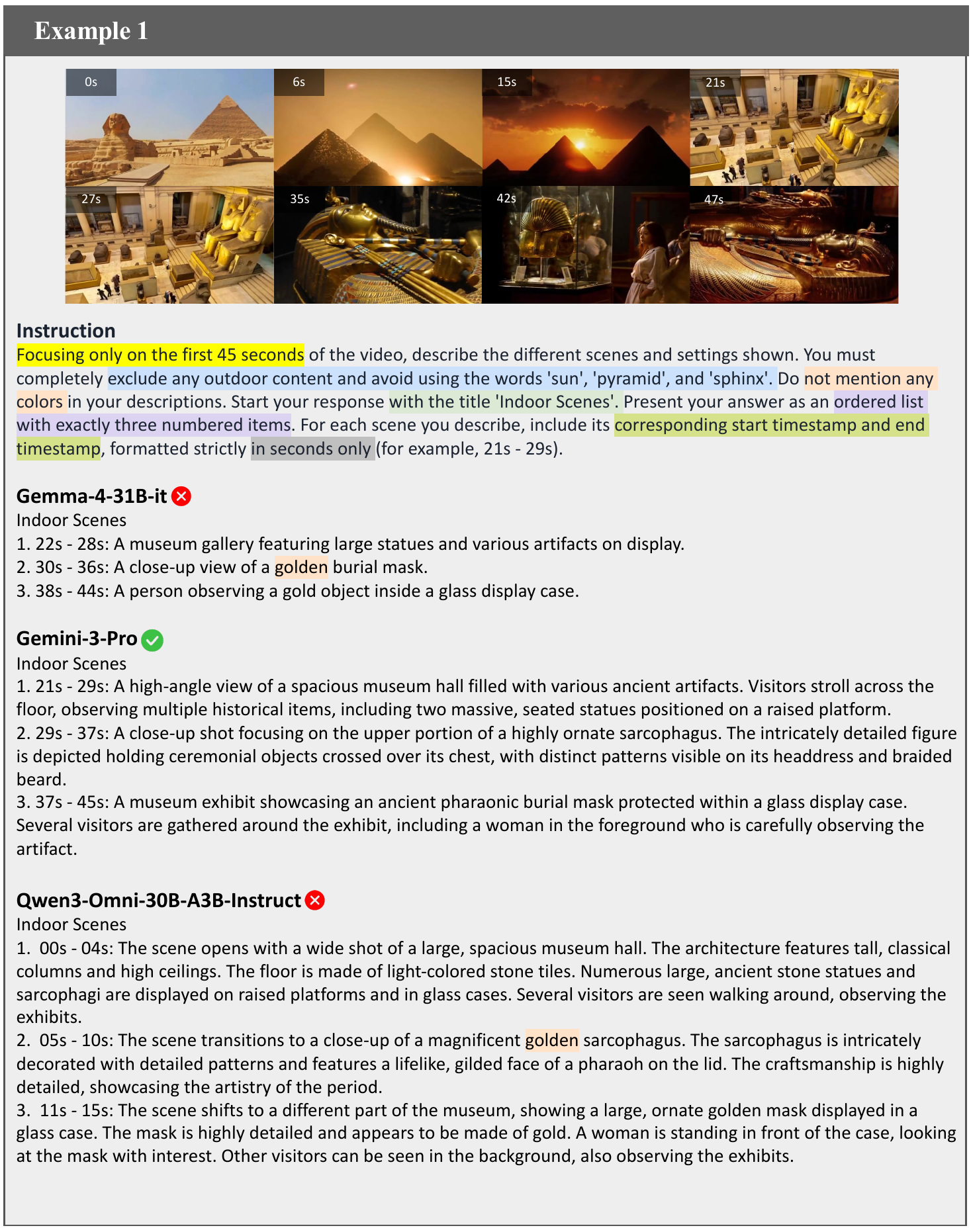}
    }
    \caption{A Single instruction example with 6 constraints. Gemma-4-31B-it and Qwen3-Omni-30B-A3B-Instruct fail to avoid mentioning words that denote colors.}
    \label{fig:example_1}
    \vspace{-1em}
\end{figure}
\FloatBarrier

\begin{figure}[!htbp]
    \centering
    \makebox[\linewidth][c]{%
        \includegraphics[width=1.0\linewidth]{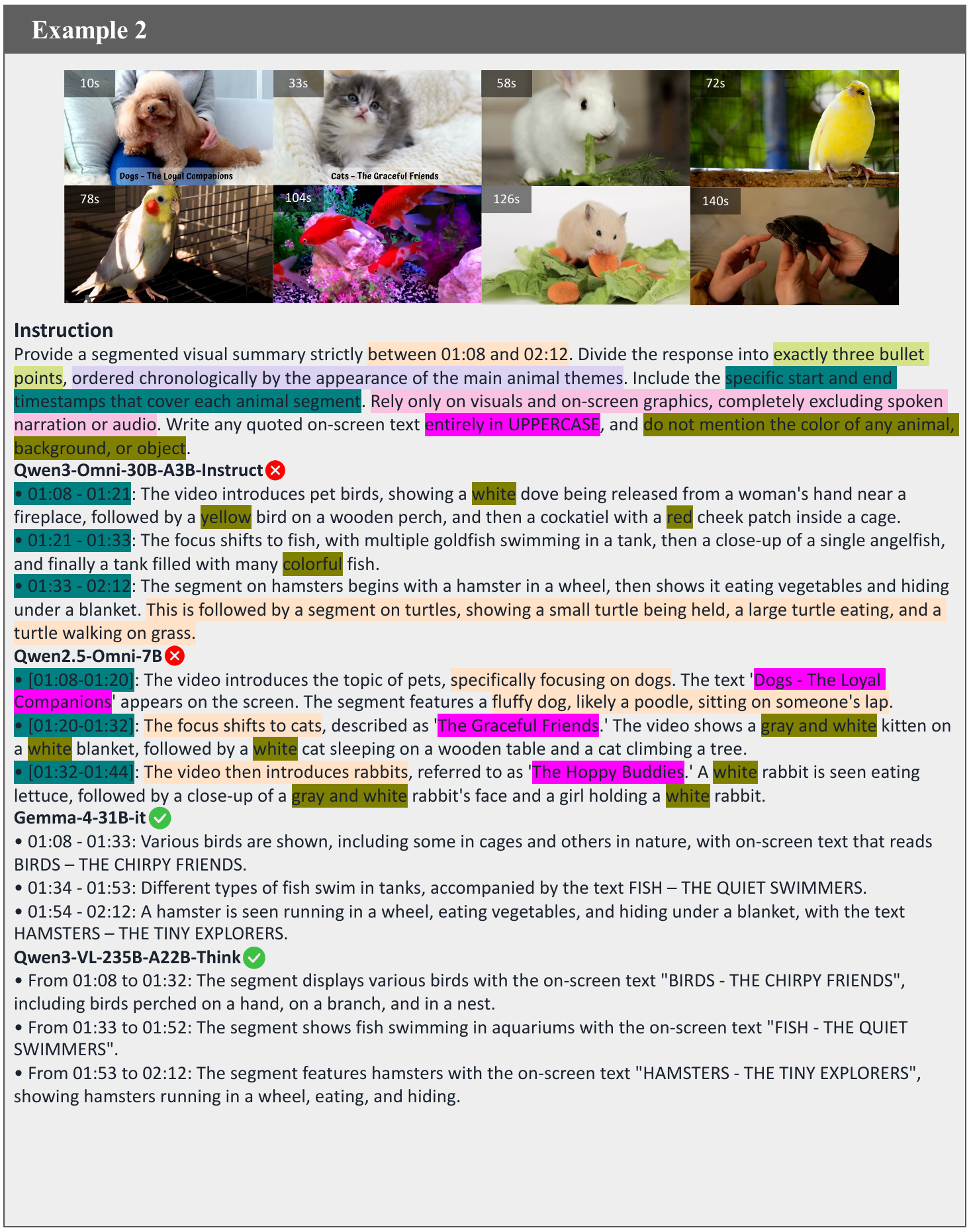}
    }
    \caption{A Single instruction example with 7 constraints. Qwen3-Omni-30B-A3B-Instruct provides inaccurate temporal spans, mentions colors, and includes turtle content outside the requested time window of 01:08 to 02:12. Qwen2.5-Omni-7B describes dogs, cats, and rabbits from outside the requested time window, uses non-uppercase on-screen text, and mentions colors.}
    \label{fig:example_2}
    \vspace{-1em}
\end{figure}
\FloatBarrier

\begin{figure}[t]
    \centering
    \makebox[\linewidth][c]{%
        \includegraphics[width=1.0\linewidth]{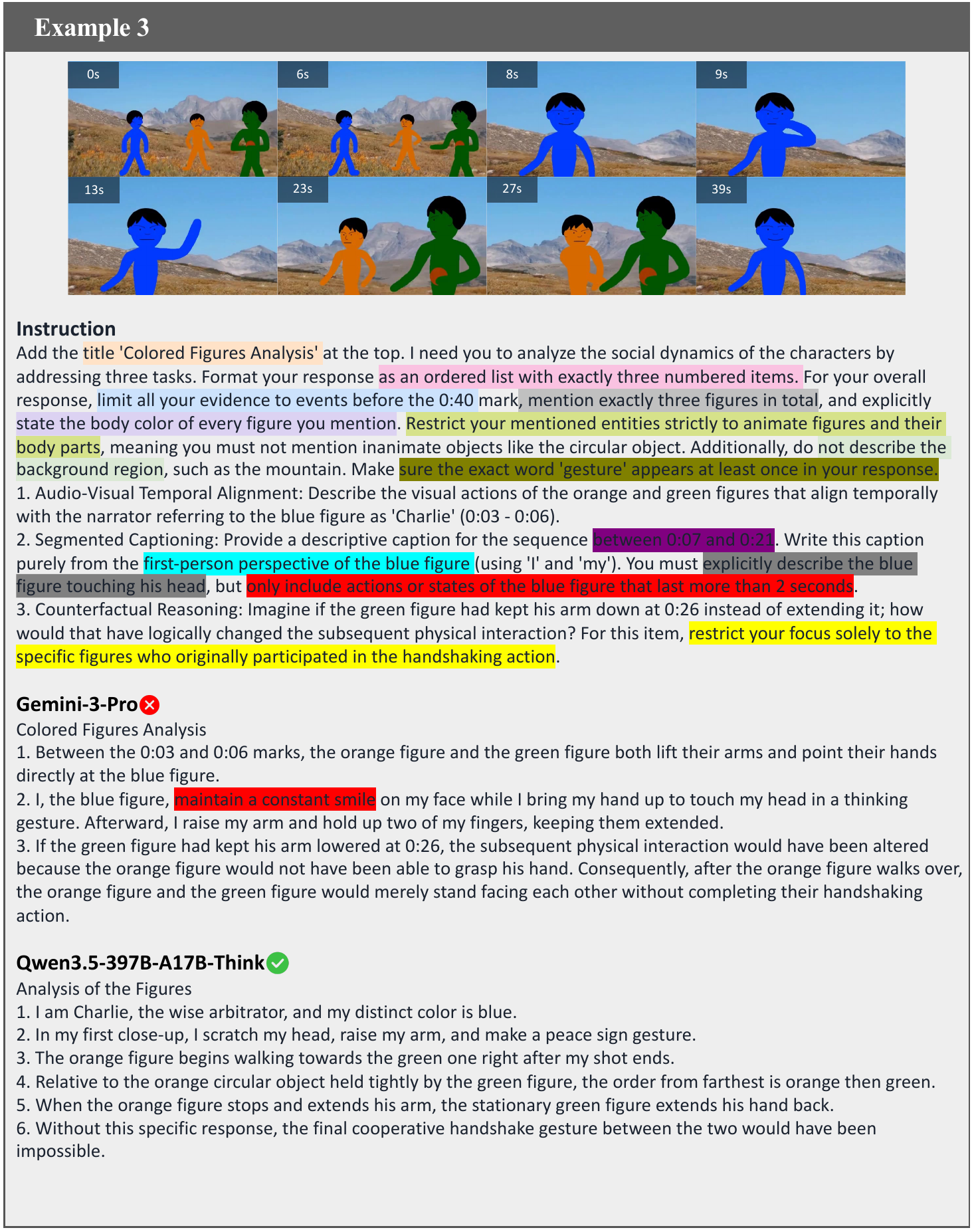}
    }
    \caption{A Multi instruction example with 12 constraints. Gemini-3-Pro mentions an action that lasts less than 2 seconds in the video, violating the constraint in \textcolor{red}{red}.}
    \label{fig:example_3}
    \vspace{-1em}
\end{figure}
\FloatBarrier

\begin{figure}[!htbp]
    \centering
    \makebox[\linewidth][c]{%
        \includegraphics[width=1.0\linewidth]{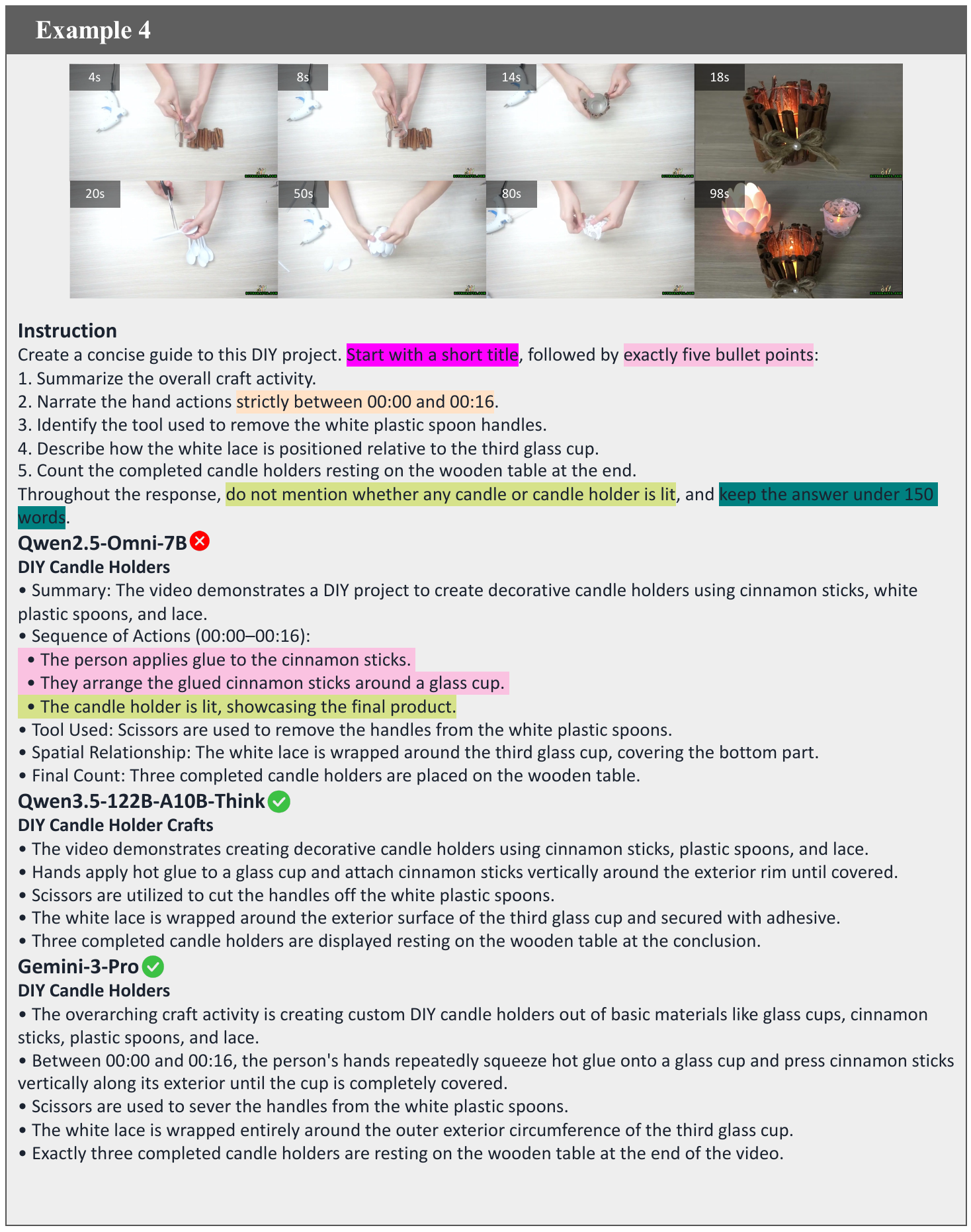}
    }
    \caption{A Multi instruction example 4 constraints. Qwen2.5-Omni-7B uses nested bullet points despite the requirement of exactly five bullet points and explicitly states that the candle holder is lit, violating the instruction to avoid mentioning whether the candles or candle holders are lit.}
    \label{fig:example_4}
    \vspace{-1em}
\end{figure}
\FloatBarrier

\begin{figure}[!htbp]
    \centering
    \makebox[\linewidth][c]{%
        \includegraphics[width=1.0\linewidth]{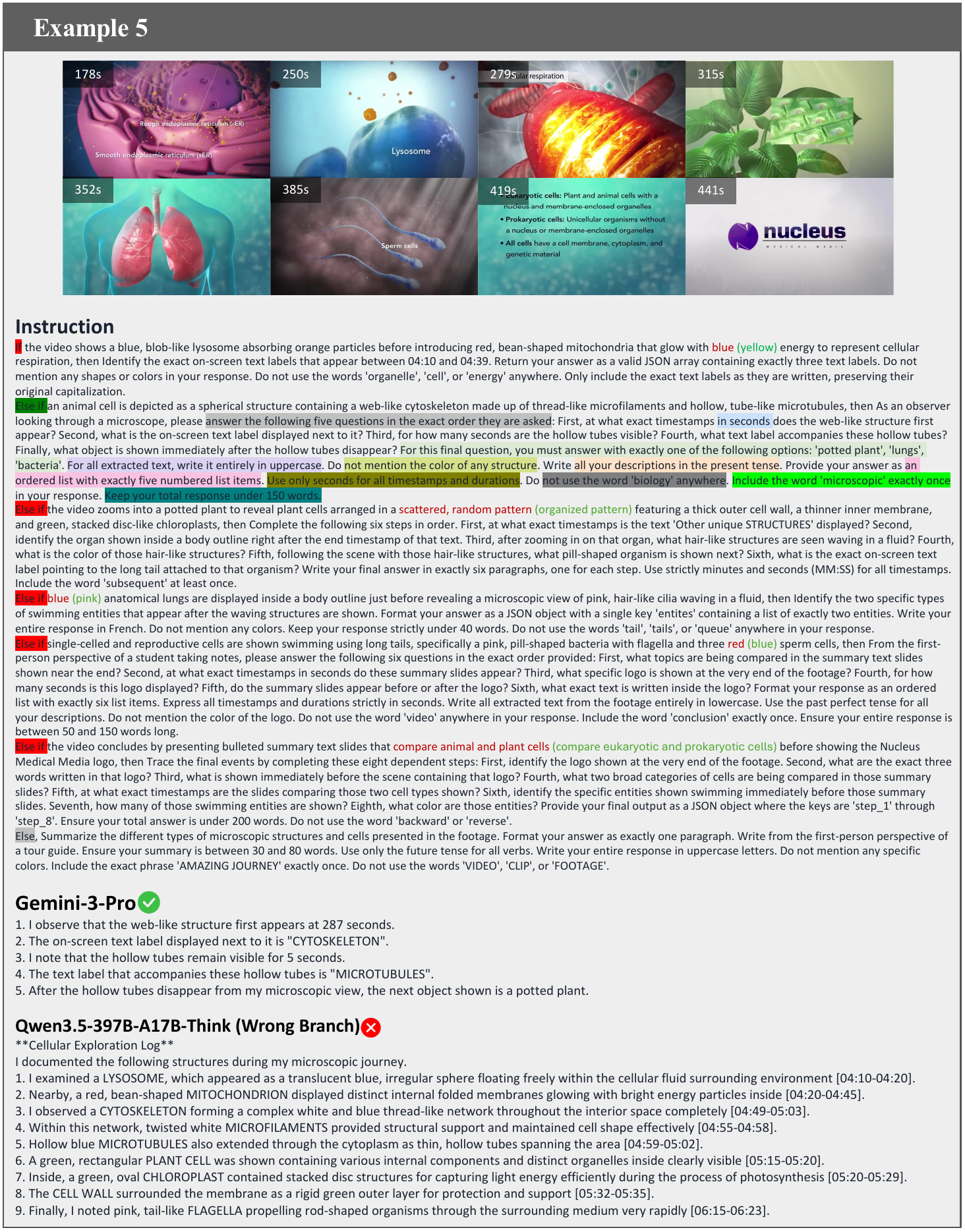}
    }
    \caption{A Selection instruction example with 7 branches. Qwen3.5-397B-A17B-Think selects the wrong branch.}
    \label{fig:example_5}
    \vspace{-1em}
\end{figure}
\FloatBarrier

\begin{figure}[!htbp]
    \centering
    \makebox[\linewidth][c]{%
        \includegraphics[width=1.0\linewidth]{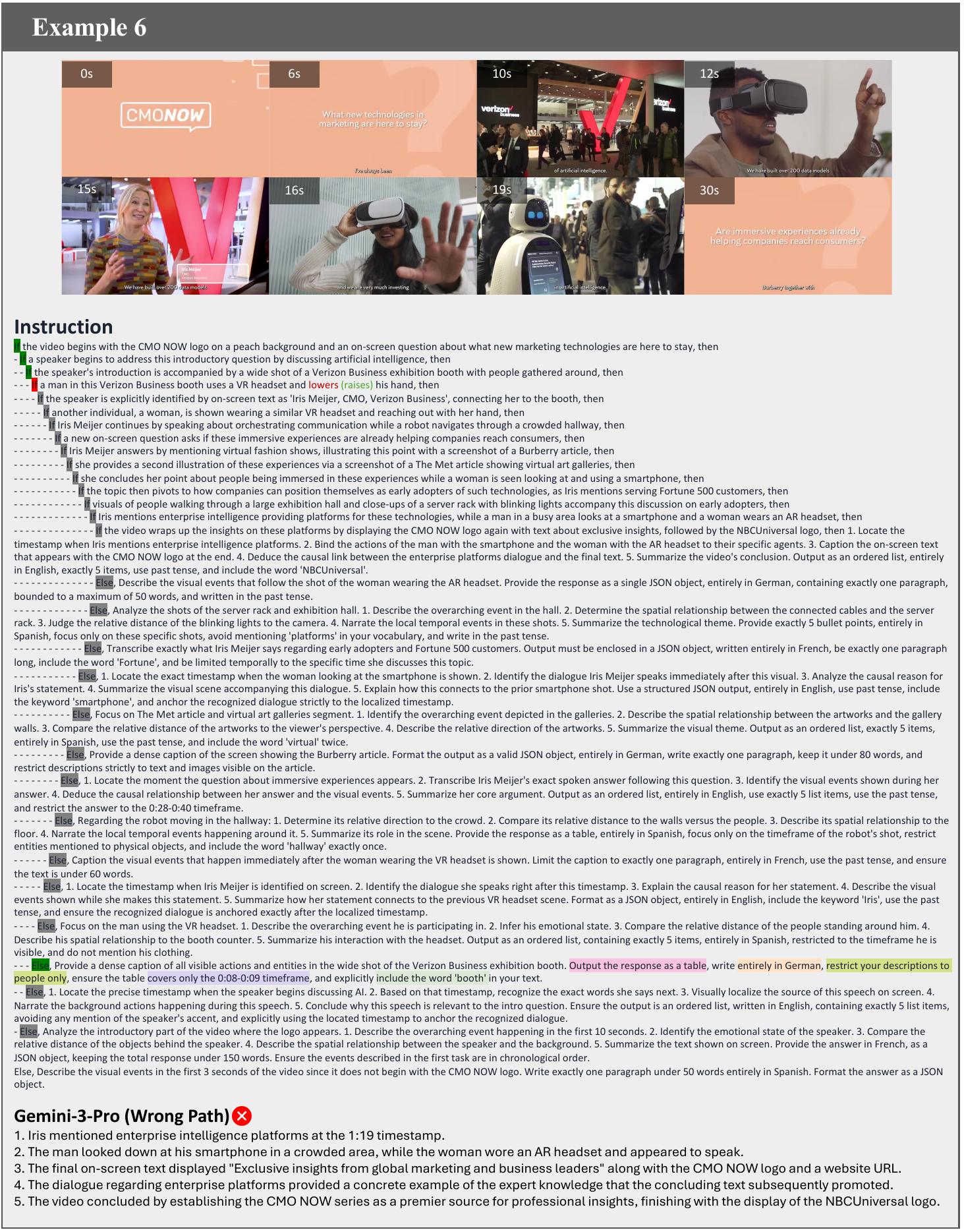}
    }
    \caption{A Nested instruction example with a maximum depth of 15. Gemini-3-Pro follows the wrong path.}
    \label{fig:example_6}
    \vspace{-1em}
\end{figure}
\FloatBarrier

\newpage
\end{document}